\newif\ifanonymous
\anonymousfalse

\documentclass[11pt]{article}
\usepackage[margin=1in]{geometry}
\usepackage{setspace}
\usepackage{amsmath,amssymb,amsfonts,mathtools,amsthm}
\usepackage{bm}
\usepackage{enumitem}
\usepackage{graphicx}
\usepackage{booktabs}
\usepackage{float}
\newfloat{algorithm}{htbp}{loa}
\floatname{algorithm}{Algorithm}
\usepackage{placeins}
\usepackage{natbib}
\bibpunct[, ]{(}{)}{,}{a}{}{,}

\usepackage{xurl}
\usepackage[colorlinks=true,linkcolor=blue,citecolor=blue,urlcolor=blue]{hyperref}
\usepackage{microtype}

\allowdisplaybreaks

\numberwithin{equation}{section}

\DeclareMathOperator*{\argmin}{arg\,min}
\newcommand{\RR}{\mathcal R}
\newcommand{\Dpax}{\mathcal D^{\mathrm{pax}}}
\newcommand{\Doff}{\mathcal D^{\mathrm{off}}}
\theoremstyle{plain}
\newtheorem{lemma}{Lemma}[section]
\newtheorem{proposition}[lemma]{Proposition}

\theoremstyle{definition}
\newtheorem{assumption}[lemma]{Assumption}

\title{Multistage Contextual Optimization for Vehicle Repositioning\\[0.4em]}
\ifanonymous
\author{}
\else
\author{
Jinyu Zhang\textsuperscript{1,3},
Di Huang\textsuperscript{1,*},
Zhiyuan Liu\textsuperscript{1,2}, and
Nikolas Geroliminis\textsuperscript{3}\\[0.6em]
\small \textsuperscript{1}School of Transportation, Southeast University, Nanjing, China\\
\small \textsuperscript{2}School of Statistics and Data Science, Southeast University, Nanjing, China\\
\small \textsuperscript{3}Urban Transport Systems Laboratory (LUTS),\\
\small \'Ecole Polytechnique F\'ed\'erale de Lausanne (EPFL), Lausanne, Switzerland\\[0.4em]
\small \textsuperscript{*}Corresponding author
}
\fi
\date{}

\begin{document}
\maketitle

\begin{abstract}
Ride-hailing platforms reposition idle vehicles to improve spatial supply--demand balances. This problem is prediction-driven and multistage: at each stage, the platform forecasts future demand, executes current decisions, and uses the resulting state to initialize the next stage. We formulate a multistage linear program (LP) with a macroscopic fundamental diagram (MFD) linking accumulation to trip completion. We then develop a multistage contextual optimization approach that uses decision performance to train the prediction model. The learning problem combines multiple coupled stages, context that depends on earlier decisions, and predictions that affect the objective, the constraint matrix, and the right-hand side of the LP. We derive the exact gradient, which has three parts: parameters change forecasts and hence decisions; earlier decisions change the current decision through the inherited state; and this state also changes forecasts as context. An equivalent adjoint uses reverse vector--Jacobian products, avoiding full solution and state-parameter Jacobians. Finally, forward propagation solves many stage LPs one after another, which makes training slow. We propose a parallel block evaluation algorithm that solves blocks of consecutive stages in parallel and repeatedly feeds each block output to the next block. We prove that the algorithm recovers the serial trajectory within a finite number of passes and that its gradient error remains bounded when it stops early. On New York City ride-hailing data, multistage contextual optimization yields 7.7\% to 9.1\% more business benefit than the same predictor trained by mean squared error, outperforms all compared baselines from five paradigms, and even outperforms realized demand fed to the same policy. With eight workers, parallel block evaluation shortens training by a factor of 1.86 without loss of decision performance.
\end{abstract}

\noindent\textbf{Keywords:} ride-hailing; vehicle repositioning; macroscopic fundamental diagram; multistage contextual optimization.

\section{Introduction}
Ride-hailing demand varies across space and time, while the vehicle distribution reflects earlier trips. Vehicles may remain idle in one region as requests wait in another \citep{ZhangEtAl2023RIVER}. Repositioning can correct this imbalance, but it incurs an immediate empty-trip cost and removes a vehicle from service until it arrives \citep{PavoneEtAl2012,PantelidisEtAl2022}. The problem is therefore prediction-driven and multistage. At each stage, the platform forecasts future demand, executes current repositioning and service under the available information, and updates the fleet state. The resulting idle and in-motion stocks become the initial state for the next stage.

Repositioning and service also interact with congestion across the city. Empty repositioning, occupied trips, and vehicles already in motion all contribute to regional accumulation \citep{XuEtAl2021RideSourcing}. Near the production-maximizing regime, an additional trip consumes scarce road capacity and delays vehicle return. We argue that in systems with a large number of ride-hailing vehicles (as in many large cities worldwide), vehicle repositioning can influence the level of congestion and regional speeds in different parts of the network. The macroscopic fundamental diagram (MFD) describes this effect at the same regional level at which repositioning is decided: it represents an aggregate, robust, and empirically validated relation between vehicle accumulation and production or speed \citep{GeroliminisDaganzo2008,PaipuriEtAl2021}. The MFD has recently been integrated into dynamic ride-sourcing models \citep{RamezaniValadkhani2023,BeojoneGeroliminis2023} and vehicle-repositioning methods \citep{ValadkhaniRamezani2023,BeojoneEtAl2024,HuangEtAl2024} to capture congestion feedback.

Ride-hailing demand predictors are usually trained for forecast accuracy, such as OD-level spatio-temporal deep learning models \citep{KeEtAl2021}. However, the cost of a forecast error depends on where idle vehicles are, and their locations depend on earlier repositioning and trips. Contextual optimization addresses this mismatch: it predicts uncertain parameters for a downstream optimization model and trains the predictor on decision performance rather than numerical forecast error \citep{BertsimasKallus2020,ElmachtoubGrigas2022,HuangEtAl2024SemiSPO,HuangEtAl2025Prescriptive}. Methodological extensions handle uncertain constraints \citep{HuEtAl2023Constraints}, decision-dependent uncertainty \citep{LiuEtAl2022Endogenous}, and multistage problems \citep{HuEtAl2024Multistage,PersakAnjos2024DFF}. The two multistage studies propagate gradients across stages, but neither captures how the inherited state changes the forecast as context, and \citet{PersakAnjos2024DFF} restrict uncertainty to the objective. In transportation, recent studies apply contextual optimization to vehicle repositioning and show that decision-based training can improve downstream performance \citep{GuoEtAl2026,WangEtAl2026SPOVCS}. However, they treat vehicle repositioning as a single optimization stage, so they cannot capture how current decisions change the state and operating cost of later stages.

Existing studies therefore leave a gap between MFD-aware repositioning and contextual optimization. MFD-aware repositioning takes demand forecasts as given, and contextual optimization has not been developed for a multistage policy in which the forecast sets the current decisions, these decisions change the next state, and this state is again an input of the predictor. Closing this gap raises three challenges. First, demand often enters operating constraints, so decisions computed from forecasts must remain feasible when current demand is realized. Second, each inherited state enters both the predictor and the operational model, so gradients must propagate through both paths and the recursive state trajectory; without the path through the predictor, the gradient misses how the parameters change later forecasts. Third, training repeatedly solves optimization models across many stages. Existing acceleration methods reduce the cost of differentiation \citep{AmosKolter2017,BerthetEtAl2020} or the number of solver calls \citep{MulambaEtAl2021}, and \citet{HuEtAl2024Multistage} update per-stage networks in parallel, but none of them shortens the sequential forward evaluation of one sample, which remains the bottleneck of training.

To close this gap, this paper makes three contributions. First, we formulate MFD-aware multistage vehicle repositioning as a rolling commitment--recourse policy. It separates pre-realization repositioning from post-realization service. We linearize the MFD dynamics and penalize an upper bound on the resulting movement error. Second, we develop a multistage contextual optimization approach in which the predictor uses the decision-induced operating state as context. Our problem involves coupled stages, decision-dependent context, and forecasts that change the objective, constraint matrix, and right-hand sides (RHS) of every stage linear program (LP). For this problem, we derive the exact gradient of the complete multistage graph, together with an equivalent adjoint that avoids full solution and state-parameter Jacobians. Third, we propose a parallel block evaluation algorithm that solves blocks of stages in parallel to accelerate training. We prove that it recovers the serial trajectory within finitely many passes and that its gradient error stays bounded under early stopping.

\section{MFD-Aware Rolling-Horizon Repositioning Model}
\label{sec:model}

This section constructs an MFD-aware vehicle-repositioning model. We first describe the problem setting (Section~\ref{sec:problem-setting}) and formulate the multi-period planning problem (Section~\ref{sec:multi-period-model}). We then derive an affine approximation of its traffic dynamics (Section~\ref{sec:period-level-pwl-mfd-lp}) and describe multistage execution (Section~\ref{sec:multi-stage-rolling-horizon}). Finally, we make decisions under forecast demand with a commitment--recourse scheme (Section~\ref{sec:commitment-recourse-execution}) and update its reference trajectory (Section~\ref{sec:reference-update}).

\subsection{Problem Setting}
\label{sec:problem-setting}

\begin{figure}[htbp]
  \centering
  \includegraphics[width=0.9\textwidth]{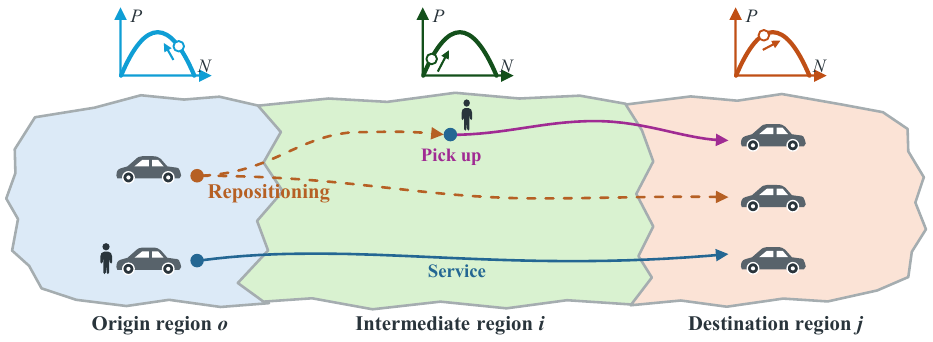}
  \caption{Vehicle movements and their effect on regional traffic. A service trip carries a passenger from origin region $o$ to destination region $j$. A repositioning vehicle travels empty from $o$ toward $j$ and may pick up a passenger in the intermediate region $i$ who is also headed to $j$. Above each region, the MFD relates production $P$ to accumulation $N$. The arrows show how these movements shift the regional state: accumulation falls in the congested origin region, which raises its production, and rises in the regions the vehicles enter.}
  \label{fig:problem-setting}
\end{figure}

We study an urban ride-hailing platform that operates across service regions. At each decision stage, its operating state consists of idle vehicles, waiting requests, and vehicles in motion. It forecasts future regional demand and proactively repositions part of the idle fleet toward regions expected to face high demand. This action can improve the future supply--demand balance, but it incurs empty-travel cost. The platform therefore coordinates service and repositioning over time to limit the waiting-request backlog, request abandonment, and empty travel. Repositioning also changes the traffic state. Idle, occupied, and empty in-motion platform vehicles contribute to regional accumulation together with background traffic. At high accumulation, a further increase can reduce production and delay vehicle return. We represent this feedback with a regional MFD. The model thus captures both effects of repositioning: it changes the future supply distribution and the congestion under which that supply becomes available. Figure~\ref{fig:problem-setting} illustrates both effects in three regions.

At each decision stage, the platform makes two linked decisions under different information. It first uses the current operating state and a rolling demand forecast to commit part of the idle fleet to repositioning. The remaining idle vehicles stay available for passenger service. After current requests are observed, the platform decides how to serve new and waiting requests subject to the earlier repositioning commitment. Repositioning and service change vehicle locations and the waiting-request backlog, while MFD dynamics determine vehicle movement. The state reached after the current period initializes the next stage. The problem is multistage because current actions affect later operating states and hence the decisions made there.

\subsection{Multi-Period Repositioning Model}
\label{sec:multi-period-model}

Let $\RR$ be the region set and $\mathcal E\subseteq\RR\times\RR$ the directed edge set connecting adjacent regions. The planning horizon contains $H\ge1$ periods of length $\Delta>0$, where $H$ is an integer, indexed by $\mathcal H=\{0,\ldots,H-1\}$. Let $\Dpax=\RR\times\RR$ be the passenger origin--destination (OD) set. Repositioning uses its off-diagonal subset $\Doff:=\{(i,j)\in\Dpax:i\ne j\}$. For $(i,j)\in\Dpax$ and $\tau\in\mathcal H$, the known input $d_{ij,\tau}$ is the number of new requests. 

\subsubsection*{Movement classes}

The platform can send a vehicle through the network in three ways. An idle vehicle can carry a passenger directly from an origin to a destination. It can instead travel empty toward a target destination as a repositioning action. During this empty trip, it may also pick up a passenger in an intermediate region and carry that passenger to the same target destination (Figure~\ref{fig:problem-setting}). This loaded pickup conversion uses the same off-diagonal OD set $\Doff$. We denote the ordinary passenger-service, empty-repositioning, and loaded pickup-conversion class sets by \(\mathcal A^{\mathrm{srv}}:=\{a_{ij}^{\mathrm{srv}}:(i,j)\in\Dpax\}\), \(\mathcal A^{\mathrm{emp}}:=\{a_{oj}^{\mathrm{emp}}:(o,j)\in\Doff\}\), and \(\mathcal A^{\mathrm{load}}:=\{a_{ij}^{\mathrm{load}}:(i,j)\in\Doff\}\), and let \(\mathcal A\) be their union. Here $o$ is the original origin of an empty itinerary, while $i$ is a passenger pickup origin. For each $(i,j)\in\Dpax$, let $\mathcal P_{ij}=(p^{ij}_0=i,p^{ij}_1,\ldots,p^{ij}_{L_{ij}}=j)$ be a fixed directed path whose consecutive regions are adjacent in $\mathcal E$. The path contains $L_{ij}+1$ region positions and $L_{ij}$ directed edges. For each $a\in\mathcal A$, let $\mathsf{o}(a)$ and $\mathsf{d}(a)$ denote its origin and destination. Class $a$ uses $\mathcal P_{\mathsf{o}(a),\mathsf{d}(a)}$, and $n^a_{k,\tau}$ is its vehicle stock in region $k$ at the start of period $\tau$. Stocks and flows outside a class's path are zero throughout the model. An empty class $a_{oj}^{\mathrm{emp}}$ may pick up an $i\to j$ passenger at an internal region $i\in\mathcal P_{oj}\setminus\{o,j\}$ when $(i,j)\in\Doff$.

\subsubsection*{Dispatch}

At the beginning of period $\tau$, $e_{k,\tau}$ is the idle stock in region $k$. Idle vehicles can enter ordinary service or repositioning. The variable $m_{ij,\tau}$ assigns idle vehicles at passenger origin $i$ to trips toward $j$. The variable $g_{oj,\tau}$ commits idle vehicles at original origin $o$ to empty repositioning toward a different target $j$. Let $\tilde e_{k,\tau}$ be the idle stock left after these actions. The dispatch balance is
\begin{equation}
  \tilde e_{k,\tau} + \sum_{j:\,(k,j)\in\Dpax}m_{kj,\tau}
  +\sum_{j:\,(k,j)\in\Doff}g_{kj,\tau}
  =e_{k,\tau},
  \qquad \forall k\in\RR,\ \tau\in\mathcal H.
  \label{eq:post-dispatch-idle}
\end{equation}
Nonnegativity of $\tilde e_{k,\tau}$ keeps total dispatch within the available idle stock. The dispatch decisions are OD-level quantities. We therefore translate them into newly created movement-class inflows. 
Let $\iota_{a,\tau}$ denote the number of newly dispatched vehicles entering class $a$ at the beginning of period $\tau$. It equals $m_{ij,\tau}$ for a service class $a_{ij}^{\mathrm{srv}}$ and $g_{oj,\tau}$ for an empty class $a_{oj}^{\mathrm{emp}}$, and it is zero for a loaded class, which receives vehicles only through pickup conversion.

The post-dispatch stock $\tilde n^a_{k,\tau}$ of class $a$ in region $k$ is
\begin{equation}
  \tilde n^a_{k,\tau}
  =n^a_{k,\tau}
  +\mathbf 1_{\{k=\mathsf{o}(a)\}}\iota_{a,\tau},
  \qquad \forall k\in\RR,\ a\in\mathcal A,\ \tau\in\mathcal H.
  \label{eq:post-dispatch-stock}
\end{equation}
The indicator ensures that newly dispatched class-$a$ vehicles are added only at the start region of their itinerary. Vehicles that were already in motion at the beginning of period $\tau$ remain in their current regional class stock.

\subsubsection*{Pickup conversion}

An empty vehicle in class $a_{oj}^{\mathrm{emp}}$ that starts period $\tau$ in an internal path region $i\in\mathcal P_{oj}\setminus\{o,j\}$ may serve available $i\to j$ requests. Let $y_{oij,\tau}$ be this pickup-conversion flow, defined as zero outside the eligible index combinations below. The available empty stock bounds the conversion:
\begin{equation}
  0\le y_{oij,\tau}\le \tilde n^{a_{oj}^{\mathrm{emp}}}_{i,\tau},
  \qquad
  \forall (o,j)\in\Doff,\ (i,j)\in\Doff,\ i\in\mathcal P_{oj}\setminus\{o,j\},\ \tau\in\mathcal H.
  \label{eq:pickup-feasibility}
\end{equation}
Let $\bar n^a_{k,\tau}$ denote the class stock after pickup conversion. The three class types satisfy
\begin{equation}
  \bar n^{a_{ij}^{\mathrm{srv}}}_{k,\tau}
  =
  \tilde n^{a_{ij}^{\mathrm{srv}}}_{k,\tau},
  \qquad \forall k\in\RR,\ (i,j)\in\Dpax,\ \tau\in\mathcal H.
  \label{eq:post-conversion-service}
\end{equation}
\begin{equation}
  \bar n^{a_{oj}^{\mathrm{emp}}}_{k,\tau}
  =
  \tilde n^{a_{oj}^{\mathrm{emp}}}_{k,\tau}
  -
  \mathbf 1_{\{(k,j)\in\Doff,\ k\in\mathcal P_{oj}\setminus\{o,j\}\}}y_{okj,\tau},
  \qquad \forall k\in\RR,\ (o,j)\in\Doff,\ \tau\in\mathcal H.
  \label{eq:post-conversion-empty}
\end{equation}
\begin{equation}
  \bar n^{a_{ij}^{\mathrm{load}}}_{k,\tau}
  =
  \tilde n^{a_{ij}^{\mathrm{load}}}_{k,\tau}
  +
  \mathbf 1_{\{k=i\}}
  \sum_{o:\,(o,j)\in\Doff,\ i\in\mathcal P_{oj}\setminus\{o,j\}}
  y_{oij,\tau},
  \qquad \forall k\in\RR,\ (i,j)\in\Doff,\ \tau\in\mathcal H.
  \label{eq:post-conversion-loaded}
\end{equation}

\subsubsection*{Demand backlog and abandonment}
Let \(K^{\mathrm{wait}}\) be the nonnegative integer number of additional periods for which a request remains available after its arrival period. For service period \(\tau\), define the admissible arrival-period set \(\mathcal T_\tau^{\mathrm{arr}}:=\{\tau-K^{\mathrm{wait}},\ldots,\tau\}\). For \(\tau'\in\mathcal T_\tau^{\mathrm{arr}}\), let \(q_{ij,\tau,\tau'}\) be the corresponding request stock available before service. Requests of one OD pair that arrived in the same period \(\tau'\) form an arrival cohort. The current cohort holds the new requests, \(q_{ij,\tau,\tau}=d_{ij,\tau}\). Ordinary service and pickup conversion can serve cohorts with \(\tau'\in\mathcal T_\tau^{\mathrm{arr}}\). Let \(\mu_{ij,\tau,\tau'}\) be the service assigned to the cohort from \(\tau'\), and let \(\mu_{ij,\tau}\) be the total volume served. Summing the cohort allocations must equal the passenger-service flow supplied by ordinary dispatch and pickup conversion:
\begin{equation}
  \mu_{ij,\tau}
  :=\sum_{\tau'\in\mathcal T_\tau^{\mathrm{arr}}}
  \mu_{ij,\tau,\tau'}
  =m_{ij,\tau}
  +\sum_{o:\,(o,j)\in\Doff,\ i\in\mathcal P_{oj}\setminus\{o,j\}}
  y_{oij,\tau},
  \qquad \forall (i,j)\in\Dpax,\ \tau\in\mathcal H.
  \label{eq:served-demand}
\end{equation}

For a nonterminal cohort, the stock left after service remains available in the next period:
\begin{equation}
  q_{ij,\tau+1,\tau'}
  =q_{ij,\tau,\tau'}-\mu_{ij,\tau,\tau'},
  \qquad
  \forall \tau'\in\mathcal T_\tau^{\mathrm{arr}}\setminus
  \{\tau-K^{\mathrm{wait}}\},
  \quad \forall (i,j)\in\Dpax,\ \tau\in\mathcal H.
  \label{eq:backlog-propagation}
\end{equation}

The cohort that arrived in \(\tau-K^{\mathrm{wait}}\) is in its final service period. Its remaining stock, denoted by \(A_{ij,\tau}\), abandons at the end of period \(\tau\):
\begin{equation}
  A_{ij,\tau}
  =q_{ij,\tau,\tau-K^{\mathrm{wait}}}
  -\mu_{ij,\tau,\tau-K^{\mathrm{wait}}},
  \qquad \forall (i,j)\in\Dpax,\ \tau\in\mathcal H.
  \label{eq:backlog-abandonment}
\end{equation}
Nonnegativity of the next-period stocks and abandonment variables ensures that service does not exceed the available stock of any cohort. Let \(q_{ij,\tau}\) denote the inherited backlog at the start of period \(\tau\), excluding new demand \(d_{ij,\tau}\). This aggregate request count is
\begin{equation}
  q_{ij,\tau}
  :=
  \sum_{\tau'\in\mathcal T_\tau^{\mathrm{arr}}\setminus\{\tau\}}
  q_{ij,\tau,\tau'},
  \qquad\forall (i,j)\in\Dpax,\ \tau=0,\ldots,H.
  \label{eq:aggregate-backlog}
\end{equation}

The initial state fixes the first-period stocks: \(e_{i,0}=e_i^0\), \(q_{ij,0,\tau'}=q_{ij,0,\tau'}^0\) for \(\tau'\in\mathcal T_0^{\mathrm{arr}}\setminus\{0\}\), and \(n^a_{k,0}=n_k^{a,0}\), where \(e_i^0\), \(q_{ij,0,\tau'}^0\), and \(n_k^{a,0}\) are the given nonnegative idle, inherited cohort, and in-motion stocks.

\subsubsection*{MFD dynamics}

The MFD dynamics describe how in-motion and idle platform stocks evolve in within-period time $\vartheta\in[0,\Delta]$. Let $n^a_{k,\tau}(\vartheta)$ and $e_{k,\tau}(\vartheta)$ be the class-$a$ and idle platform stocks in region $k$ at time $\vartheta$. Let $N^{\mathrm{bg}}_{k,\tau}(\vartheta)$ be the given background traffic accumulation. The period starts after dispatch and pickup conversion, so
\begin{equation}
  e_{k,\tau}(0^+)=\tilde e_{k,\tau},\ \forall k\in\RR,\ \tau\in\mathcal H,
  \qquad
  n^a_{k,\tau}(0^+)=\bar n^a_{k,\tau},\ \forall k\in\RR,\ a\in\mathcal A,\ \tau\in\mathcal H.
  \label{eq:mfd-initial}
\end{equation}
The regional accumulation during period $\tau$ is
\begin{equation}
  N_{k,\tau}(\vartheta)
  =N^{\mathrm{bg}}_{k,\tau}(\vartheta)
  +e_{k,\tau}(\vartheta)
  +\sum_{a\in\mathcal A}n^a_{k,\tau}(\vartheta),
  \qquad \forall k\in\RR,\ \tau\in\mathcal H,\ \vartheta\in[0,\Delta].
  \label{eq:mfd-accumulation}
\end{equation}
Once an itinerary is completed, the vehicle leaves its movement class and enters the idle stock in the completion region.
Let \(N_k^{\mathrm{jam}}>0\) be the maximum accumulation of region \(k\), and let \(P_k:[0,N_k^{\mathrm{jam}}]\to\mathbb R_{\ge0}\) denote its production MFD. At accumulation \(N\), the value \(P_k(N)\) is the total regional vehicle-distance production per unit time, and the ratio \(P_k(N)/N\) is the regional speed, taken at zero as its finite right-hand limit. Accumulation stays within \([0,N_k^{\mathrm{jam}}]\). For each OD itinerary, the coefficient \(\chi^{ij,\mathrm{edge}}_{kk'}\) indicates whether \((k,k')\) is a consecutive path edge, while \(\chi^{ij,\mathrm{term}}_k\) indicates whether \(k\) is its terminal region. We define
\begin{equation}
  \begin{aligned}
  \chi^{ij,\mathrm{edge}}_{kk'}
  &:=
  \mathbf 1_{\{\exists r:\,0\le r<L_{ij},\ (k,k')=(p^{ij}_r,p^{ij}_{r+1})\}},\forall (k,k')\in\mathcal E,\ (i,j)\in\Dpax,\\
  \chi^{ij,\mathrm{term}}_k
  &:=
  \mathbf 1_{\{k=j\}},
  \qquad\forall k\in\RR,\ (i,j)\in\Dpax.
  \end{aligned}
  \label{eq:path-indicators}
\end{equation}
Let $M^a_{kk',\tau}(\vartheta)$ be the instantaneous rate at which class-$a$ vehicles transfer from region $k$ to its next region $k'$, and let $C^a_{k,\tau}(\vartheta)$ be the instantaneous rate at which class-$a$ vehicles complete their path in region $k$. Let \(\bar\ell_k>0\) be the average distance of one regional vehicle movement in region \(k\). Dividing production by this distance gives the aggregate regional outflow rate \(P_k(N)/\bar\ell_k\), measured in vehicles per unit time. Each movement class receives a share proportional to its stock relative to total regional accumulation. At zero accumulation, both rates are defined as zero:
\begin{equation}
  M^a_{kk',\tau}(\vartheta)
  =
  \chi^{\mathsf{o}(a)\mathsf{d}(a),\mathrm{edge}}_{kk'}
  \frac{n^a_{k,\tau}(\vartheta)}{N_{k,\tau}(\vartheta)}
  \frac{P_k(N_{k,\tau}(\vartheta))}
  {\bar\ell_k},
  \qquad \forall (k,k')\in\mathcal E,\ a\in\mathcal A,\ \tau\in\mathcal H,\ \vartheta\in[0,\Delta],
  \label{eq:mfd-transfer-rate}
\end{equation}
\begin{equation}
  C^a_{k,\tau}(\vartheta)
  =
  \chi^{\mathsf{o}(a)\mathsf{d}(a),\mathrm{term}}_k
  \frac{n^a_{k,\tau}(\vartheta)}{N_{k,\tau}(\vartheta)}
  \frac{P_k(N_{k,\tau}(\vartheta))}
  {\bar\ell_k},
  \qquad \forall k\in\RR,\ a\in\mathcal A,\ \tau\in\mathcal H,\ \vartheta\in[0,\Delta].
  \label{eq:mfd-completion-rate}
\end{equation}
The stock of class-$a$ vehicles and the idle stock in region $k$ evolve according to the coupled nonlinear ordinary differential equations (ODEs)
\begin{equation}
  \begin{aligned}
  \frac{\mathrm d}{\mathrm d\vartheta}n^a_{k,\tau}(\vartheta)
  ={}&
  \sum_{k':(k',k)\in\mathcal E}M^a_{k'k,\tau}(\vartheta)
  -\sum_{k':(k,k')\in\mathcal E}M^a_{kk',\tau}(\vartheta)
  -C^a_{k,\tau}(\vartheta),\forall k\in\RR,\ a\in\mathcal A,\ \tau\in\mathcal H,\ \vartheta\in[0,\Delta],\\
  \frac{\mathrm d}{\mathrm d\vartheta}e_{k,\tau}(\vartheta)
  ={}&
  \sum_{a\in\mathcal A:\,\mathsf{d}(a)=k}C^a_{k,\tau}(\vartheta),
  \qquad\forall k\in\RR,\ \tau\in\mathcal H,\ \vartheta\in[0,\Delta].
  \end{aligned}
  \label{eq:mfd-odes}
\end{equation}

Let $\bar M^a_{kk',\tau}$ and $\bar C^a_{k,\tau}$ denote the integrated period transfer and completion volumes:
\begin{equation}
  \begin{aligned}
  \bar M^a_{kk',\tau}
  &=
  \int_0^\Delta
  M^a_{kk',\tau}(\vartheta)\,\mathrm d\vartheta,
  &&\forall (k,k')\in\mathcal E,\ a\in\mathcal A,\ \tau\in\mathcal H,\\
  \bar C^a_{k,\tau}
  &=
  \int_0^\Delta
  C^a_{k,\tau}(\vartheta)\,\mathrm d\vartheta,
  &&\forall k\in\RR,\ a\in\mathcal A,\ \tau\in\mathcal H.
  \end{aligned}
  \label{eq:mfd-integrated-volumes}
\end{equation}
The end-of-period state transition is
\begin{equation}
  \begin{aligned}
  n^a_{k,\tau+1}
  &=
  \bar n^a_{k,\tau}
  +\sum_{k':(k',k)\in\mathcal E}\bar M^a_{k'k,\tau}
  -\sum_{k':(k,k')\in\mathcal E}\bar M^a_{kk',\tau}
  -\bar C^a_{k,\tau},
  &&\forall k\in\RR,\ a\in\mathcal A,\ \tau\in\mathcal H,\\
  e_{i,\tau+1}
  &=
  \tilde e_{i,\tau}
  +\sum_{a\in\mathcal A:\,\mathsf{d}(a)=i}\bar C^a_{i,\tau},
  &&\forall i\in\RR,\ \tau\in\mathcal H.
  \end{aligned}
  \label{eq:mfd-state-transition}
\end{equation}

\subsubsection*{Objective}

Let $c^q\ge0$ and $c^A\ge0$ be the penalties per waiting request-period and per abandoned request, common to all OD pairs. The cost $c^g_{oj}\ge0$ applies to each vehicle committed to repositioning on $(o,j)\in\Doff$. The reward $b_{ij}\ge0$ applies to each completed passenger trip on $(i,j)\in\Dpax$, whether through ordinary service or pickup conversion. The planning-horizon objective $J^{\mathrm{plan}}$ is
\begin{equation}
  \begin{aligned}
  J^{\mathrm{plan}}
  =
  \sum_{\tau=0}^{H-1}
  \left[
  \sum_{(i,j)\in\Dpax}c^q q_{ij,\tau+1}
  +\sum_{(i,j)\in\Dpax}c^A A_{ij,\tau}
  +\sum_{(o,j)\in\Doff}c^g_{oj}g_{oj,\tau}
  \right.\\
  \left.
  -\sum_{(i,j)\in\Dpax}b_{ij}\bar C^{a_{ij}^{\mathrm{srv}}}_{j,\tau}
  -\sum_{(i,j)\in\Doff}b_{ij}\bar C^{a_{ij}^{\mathrm{load}}}_{j,\tau}
  \right].
  \end{aligned}
  \label{eq:objective-original-mfd}
\end{equation}
The planning horizon ends after period $H-1$. Passengers still on board at that time complete their trips after the horizon, so \eqref{eq:objective-original-mfd} gives these trips no reward, and the plan may undervalue pickups late in the horizon. 
As in the terminal cost of finite-horizon dynamic programming, we approximate this cost-to-go by crediting occupied in-motion vehicles at the end of the horizon, whose rewards no longer depend on later decisions or demand:
\begin{equation}
  J^{\mathrm{term}}
  =
  -\nu\sum_{a\in\mathcal A^{\mathrm{srv}}\cup\mathcal A^{\mathrm{load}}}
  \sum_{k\in\RR}
  b_{\mathsf{o}(a)\mathsf{d}(a)}\,n^a_{k,H},
  \label{eq:terminal-cost}
\end{equation}
where $\nu\ge0$ is the credit fraction. The continuous-time MFD planning problem is
\begin{equation}\label{eq:base-mp-original}
  \begin{aligned}
  \min_{\substack{
  m,g,y,\mu,A,\iota,\tilde e,\\
  \tilde n,\bar n,e,q,n,N,M,C,\bar M,\bar C
  }}\quad
  J^{\mathrm{plan}}+J^{\mathrm{term}}
  \end{aligned}
\end{equation}
subject to the dispatch and pickup constraints \eqref{eq:post-dispatch-idle}--\eqref{eq:post-conversion-loaded}, demand constraints \eqref{eq:served-demand}--\eqref{eq:aggregate-backlog} with the current cohort and initial state, and MFD dynamics and state balances \eqref{eq:mfd-initial}--\eqref{eq:mfd-state-transition}.
Although vehicles and requests are discrete, we use a continuous relaxation of the allocation and state variables for regional planning. All decision variables are continuous and nonnegative.

\subsection{Affine MFD Approximation}
\label{sec:period-level-pwl-mfd-lp}

The model above links repositioning to future vehicle availability through continuous traffic dynamics. Solving it requires tracking stocks and flows within every decision period, while the MFD makes movement rates nonlinear in class stocks and regional accumulation. These features complicate the multi-period optimization. We therefore discretize the traffic dynamics and approximate the nonlinear movement equations by affine relations, yielding an LP for horizon planning.

\subsubsection*{Forward Euler Discretization}

We first approximate traffic evolution with one forward Euler step of length $\Delta$ per period. Movement rates are evaluated at the period's initial state, after dispatch and pickup conversion. The available moving and idle stocks are therefore $\bar n^a_{k,\tau}$ and $\tilde e_{k,\tau}$. Let $N^{\mathrm{bg}}_{k,\tau}=N^{\mathrm{bg}}_{k,\tau}(0)$ denote the given background accumulation at the start of the period. Total accumulation can be written as
\begin{equation}
  N_{k,\tau}=N^{\mathrm{bg}}_{k,\tau}
  +\tilde e_{k,\tau}+\sum_{a\in\mathcal A}\bar n^a_{k,\tau},
  \qquad 0\le N_{k,\tau}\le N_k^{\mathrm{jam}},
  \quad \forall k\in\RR,\ \tau\in\mathcal H.
  \label{eq:euler-accumulation}
\end{equation}

The MFD rates in \eqref{eq:mfd-transfer-rate}--\eqref{eq:mfd-completion-rate} give the following period movement fraction under the Euler approximation:
\begin{equation}
  \varphi_k(N)=\frac{\Delta P_k(N)}{\bar\ell_kN},
  \qquad \forall k\in\RR,\quad 0<N\le N_k^{\mathrm{jam}}.
  \label{eq:period-movement-fraction}
\end{equation}
At $N=0$, $\varphi_k$ is defined by continuity. Applying this fraction to the available class stock gives the period movement volume
\begin{equation}
  f^a_{k,\tau}
  =\bar n^a_{k,\tau}\varphi_k(N_{k,\tau}),
  \qquad \forall a\in\mathcal A,\ k\in\RR,\ \tau\in\mathcal H.
  \label{eq:nonlinear-period-event}
\end{equation}
We choose a sufficiently small period length $\Delta>0$ so that $0\le\varphi_k(N)<1$ for every region and all $N\in[0,N_k^{\mathrm{jam}}]$. This keeps period outflow within the available stock. Using the path indicators in \eqref{eq:path-indicators}, the Euler approximations of the integrated flows in \eqref{eq:mfd-integrated-volumes} are
\begin{equation}
  \begin{aligned}
    \bar M^a_{kk',\tau}
    &=\chi^{\mathsf{o}(a)\mathsf{d}(a),\mathrm{edge}}_{kk'}f^a_{k,\tau},
    &&\forall (k,k')\in\mathcal E,\ a\in\mathcal A,\ \tau\in\mathcal H,\\
    \bar C^a_{k,\tau}
    &=\chi^{\mathsf{o}(a)\mathsf{d}(a),\mathrm{term}}_k f^a_{k,\tau},
    &&\forall k\in\RR,\ a\in\mathcal A,\ \tau\in\mathcal H.
  \end{aligned}
  \label{eq:period-path-flows}
\end{equation}
With these flows, the period balances \eqref{eq:mfd-state-transition} replace the continuous evolution in \eqref{eq:mfd-odes}.

\subsubsection*{Affine movement equations}

The period balances are linear, but the movement volume \eqref{eq:nonlinear-period-event} still depends nonlinearly on class stock and regional accumulation. In the first period, dispatch and pickup conversion preserve total regional stock, so \(N_{k,0}\) is fixed by the initial state and \(f^a_{k,0}=\varphi_k(N_{k,0})\,\bar n^a_{k,0}\) is already linear. For later periods, we use a first-order expansion around a reference trajectory $\boldsymbol n=(\bar n^{\mathrm{ref}},N^{\mathrm{ref}})$, where $\bar n^{a,\mathrm{ref}}_{k,\tau}$ and $N^{\mathrm{ref}}_{k,\tau}$ are the reference post-conversion class stock and total regional accumulation, respectively:
\begin{equation}
  \begin{aligned}
    f^a_{k,\tau}
    &=\bar n^{a,\mathrm{ref}}_{k,\tau}\varphi_k(N^{\mathrm{ref}}_{k,\tau})
      +\varphi_k(N^{\mathrm{ref}}_{k,\tau})
       (\bar n^a_{k,\tau}-\bar n^{a,\mathrm{ref}}_{k,\tau})
      +\bar n^{a,\mathrm{ref}}_{k,\tau}\varphi'_k(N^{\mathrm{ref}}_{k,\tau})
       (N_{k,\tau}-N^{\mathrm{ref}}_{k,\tau})\\
    &=\underbrace{\varphi_k(N^{\mathrm{ref}}_{k,\tau})\bar n^a_{k,\tau}}
        _{\text{stock effect}}
      +\underbrace{\bar n^{a,\mathrm{ref}}_{k,\tau}\varphi'_k(N^{\mathrm{ref}}_{k,\tau})
        (N_{k,\tau}-N^{\mathrm{ref}}_{k,\tau})}
        _{\text{congestion effect}},
    \quad \forall a\in\mathcal A,\ k\in\RR,\ \tau\in\mathcal H\setminus\{0\}.
  \end{aligned}
  \label{eq:affine-period-event}
\end{equation}
Section~\ref{sec:reference-update} describes how this trajectory is constructed. All coefficients are evaluated before the LP is solved. We also impose the physical bounds
\begin{equation}
  0\le f^a_{k,\tau}\le\bar n^a_{k,\tau},
  \qquad \forall a\in\mathcal A,\ k\in\RR,\ \tau\in\mathcal H.
  \label{eq:period-event-stock-bound}
\end{equation}

\subsubsection*{Approximation-error control}

The affine relation matches movement at the reference, but its error can grow as planned stocks and accumulation move away from that trajectory. We seek to penalize this movement error. Its exact absolute value contains the nonlinear product $\bar n^a_{k,\tau}\varphi_k(N_{k,\tau})$, so including it directly in the objective would lose the LP structure. We first bound the error by a weighted sum of squared stock and accumulation deviations.

\begin{proposition}[Affine movement error bound]
\label{prop:affine-movement-error}
Suppose $\varphi_k$ is twice continuously differentiable on $[0,N_k^{\mathrm{jam}}]$, with $|\varphi'_k|\le \bar\varphi'_k$ and $|\varphi''_k|\le \bar\varphi''_k$ for constants $\bar\varphi'_k,\bar\varphi''_k\ge0$. The error satisfies
\begin{equation}
  \begin{gathered}
    \left|\bar n^a_{k,\tau}\varphi_k(N_{k,\tau})-f^a_{k,\tau}\right|
    \le\frac{\bar\varphi'_k}{2}\left(\bar n^a_{k,\tau}-\bar n^{a,\mathrm{ref}}_{k,\tau}\right)^2
      +\frac{\bar\varphi'_k+\bar n^{a,\mathrm{ref}}_{k,\tau}\bar\varphi''_k}{2}
       \left(N_{k,\tau}-N^{\mathrm{ref}}_{k,\tau}\right)^2,\\
    \forall a\in\mathcal A,\ k\in\mathcal P_{\mathsf{o}(a),\mathsf{d}(a)},\
    \tau\in\mathcal H\setminus\{0\}.
  \end{gathered}
  \label{eq:quadratic-movement-error-bound}
\end{equation}
\end{proposition}

Appendix~\ref{app:affine-movement-error} proves this bound. Directly penalizing the squared deviations would give a quadratic objective. To keep the planning problem linear, we replace each square by a convex piecewise-linear upper envelope.
Let $\delta$ denote either the stock deviation $\bar n^a_{k,\tau}-\bar n^{a,\mathrm{ref}}_{k,\tau}$ or the accumulation deviation $N_{k,\tau}-N^{\mathrm{ref}}_{k,\tau}$. Choose $S+1$ fixed points $(\xi_h,\xi_h^2)$ on the curve $\delta^2$, where the integer $S\ge2$ is the number of segments and $\xi_0<\cdots<\xi_S$. Set the endpoints to $\xi_0=-\max_{k\in\RR}N_k^{\mathrm{jam}}$ and $\xi_S=\max_{k\in\RR}N_k^{\mathrm{jam}}$, and include $(0,0)$ among the points. Connecting adjacent points defines the envelope $\Psi$. Each segment has slope $\gamma_h$ and intercept $\beta_h$:
\begin{equation}
  \begin{aligned}
    \Psi(\delta)
      =\underbrace{(\xi_h+\xi_{h+1})}_{\gamma_h}\delta
       +\underbrace{(-\xi_h\xi_{h+1})}_{\beta_h},\qquad\forall\delta\in[\xi_h,\xi_{h+1}],\ h\in\{0,\ldots,S-1\}.
  \end{aligned}
  \label{eq:penalty-segment}
\end{equation}
Since $\Psi$ is a maximum of finitely many affine functions, we represent the stock and accumulation envelopes by auxiliary variables $\psi^{a,\mathrm{stk}}_{k,\tau}$ and $\psi^{\mathrm{acc}}_{k,\tau}$ with linear constraints
\begin{equation}
  \begin{aligned}
    \psi^{a,\mathrm{stk}}_{k,\tau}
    &\ge\gamma_h
        (\bar n^a_{k,\tau}-\bar n^{a,\mathrm{ref}}_{k,\tau})
        +\beta_h,\forall a\in\mathcal A,\ k\in\mathcal P_{\mathsf{o}(a),\mathsf{d}(a)},\
      h\in\{0,\ldots,S-1\},\ \tau\in\mathcal H\setminus\{0\},\\
    \psi^{\mathrm{acc}}_{k,\tau}
    &\ge\gamma_h
        (N_{k,\tau}-N^{\mathrm{ref}}_{k,\tau})
        +\beta_h,\forall k\in\RR,\ h\in\{0,\ldots,S-1\},\ \tau\in\mathcal H\setminus\{0\}.
  \end{aligned}
  \label{eq:penalty-square-epigraph}
\end{equation}
The coefficients in \eqref{eq:quadratic-movement-error-bound} are nonnegative, so replacing the squares by these variables preserves the upper bound on movement error. This gives the penalty
\begin{equation}
  \begin{gathered}
    \widehat R^a_{k,\tau}
    :=\frac{\bar\varphi'_k}{2}\psi^{a,\mathrm{stk}}_{k,\tau}
      +\frac{\bar\varphi'_k+\bar n^{a,\mathrm{ref}}_{k,\tau}\bar\varphi''_k}{2}
       \psi^{\mathrm{acc}}_{k,\tau},
    \qquad\forall a\in\mathcal A,\ k\in\mathcal P_{\mathsf{o}(a),\mathsf{d}(a)},\
    \tau\in\mathcal H\setminus\{0\}.
  \end{gathered}
  \label{eq:linear-error-penalty}
\end{equation}
The penalty $\widehat R$ is the mean of $\widehat R^a_{k,\tau}$ over all index combinations in \eqref{eq:linear-error-penalty}. With weight $c^{\mathrm{mfd}}>0$, initial state $s$, demand $d$, and reference $\boldsymbol n$, the planning LP is
\begin{equation}
  \min_{\substack{
    m,g,y,\mu,A,\iota,\tilde e,\tilde n,\bar n,\\
    e,q,n,N,f,\bar M,\bar C,\psi^{\mathrm{stk}},\psi^{\mathrm{acc}}
  }}\quad
  J^{\mathrm{plan}}+J^{\mathrm{term}}+c^{\mathrm{mfd}}\widehat R,
  \label{eq:lp-mp}
\end{equation}
subject to the dispatch and pickup constraints \eqref{eq:post-dispatch-idle}--\eqref{eq:post-conversion-loaded}, demand constraints \eqref{eq:served-demand}--\eqref{eq:aggregate-backlog} with the current cohort and initial state, accumulation bounds \eqref{eq:euler-accumulation}, path flows \eqref{eq:period-path-flows}, state balances \eqref{eq:mfd-state-transition}, movement equations and stock bounds \eqref{eq:affine-period-event}--\eqref{eq:period-event-stock-bound}, and penalty constraints \eqref{eq:penalty-square-epigraph}. Stocks and flows outside class paths are zero, and all variables are continuous and nonnegative.

When platform traffic penetration is low, dispatch decisions may have little effect on regional speeds. Speeds can then be treated as exogenous, so a fixed fraction \(\min\{1,\Delta v_{k,\tau}/\bar\ell_k\}\) of each class stock in region \(k\) moves on during period \(\tau\), where \(v_{k,\tau}\) is the exogenous regional speed. These fixed-fraction movements replace the accumulation and movement equations and the penalty in \eqref{eq:lp-mp}, which gives the fixed-fraction LP.

\subsection{Multistage Repositioning}
\label{sec:multi-stage-rolling-horizon}

We link successive LP horizons through their first-period end states. Let $\mathcal T=\{0,\ldots,T-1\}$ index the $T\ge1$ decision stages. At stage $t\in\mathcal T$, $t+\ell\mid t$ denotes period $t+\ell$ planned at stage $t$, with lead $\ell\in\mathcal H$. Thus LP period $\tau\in\mathcal H$ of the stage-$t$ problem is absolute period $t+\tau$. The horizon demand vector $d_{t:t+H-1\mid t}$ stacks these periods' OD demand inputs. The multistage procedure starts from $s_0=(e^0,q^0,n^0)$:
\begin{equation}
  \begin{aligned}
    e_{i,0\mid 0}&=e_i^0,
    &&\forall i\in\RR,\\
    q_{ij,0,\tau'\mid 0}&=q_{ij,0,\tau'}^0,
    &&\forall (i,j)\in\Dpax,\ \tau'\in\mathcal T_0^{\mathrm{arr}}\setminus\{0\},\\
    n^a_{k,0\mid 0}&=n_k^{a,0},
    &&\forall k\in\RR,\ a\in\mathcal A.
  \end{aligned}
  \label{eq:rolling-initial-state}
\end{equation}

At each stage $t\in\mathcal T$, solve \eqref{eq:lp-mp} from $s_t$ with horizon demand $d_{t:t+H-1\mid t}$ and a given reference trajectory. Let $u$ collect its decision variables and $u_t^\star$ denote an optimal plan. Only current-period actions are executed. Let $z_t$ collect these service, repositioning, and pickup decisions together with the first-period backlog, abandonment, and trip completions used to evaluate cost. Let $\Pi_z$ be their fixed selector:
\begin{equation}
  z_t=\Pi_z u_t^\star,\qquad \forall t\in\mathcal T.
	\label{eq:executed-block}
\end{equation}
The first-period end state in this optimal plan directly initializes the next planning horizon:
\begin{equation}
  \begin{aligned}
    e_{i,t+1\mid t+1}&=e^\star_{i,t+1\mid t},
    &&\forall i\in\RR,\ t\in\mathcal T\setminus\{T-1\},\\
    q_{ij,t+1,\tau'\mid t+1}&=q^\star_{ij,t+1,\tau'\mid t},
    &&\forall (i,j)\in\Dpax,\ \tau'\in\mathcal T_{t+1}^{\mathrm{arr}}\setminus\{t+1\},\ t\in\mathcal T\setminus\{T-1\},\\
    n^a_{k,t+1\mid t+1}&=n^{a,\star}_{k,t+1\mid t},
    &&\forall k\in\RR,\ a\in\mathcal A,\ t\in\mathcal T\setminus\{T-1\}.
  \end{aligned}
  \label{eq:rolling-state-assignment}
\end{equation}
Let $\Pi_s$ select these stocks from the plan. Then $s_{t+1}=\Pi_s u_t^\star$ for all $t\in\mathcal T$. We evaluate each stage by the operating cost of its executed decisions. The coefficient vector $c$ assigns the costs in \eqref{eq:objective-original-mfd} to the components of $z_t$:
\begin{equation}
  \begin{aligned}
    J_t^{\mathrm{exec}}=c^\top z_t
    ={}&c^q \sum_{(i,j)\in\Dpax} q^\star_{ij,t+1\mid t}
       +c^A\sum_{(i,j)\in\Dpax} A^\star_{ij,t\mid t}
       +\sum_{(o,j)\in\Doff}c^g_{oj}g^\star_{oj,t\mid t}\\
       &-\sum_{(i,j)\in\Dpax}b_{ij}\bar C^{a_{ij}^{\mathrm{srv}},\star}_{j,t\mid t}
       -\sum_{(i,j)\in\Doff}b_{ij}\bar C^{a_{ij}^{\mathrm{load}},\star}_{j,t\mid t},\qquad \forall t\in\mathcal T.
  \end{aligned}
  \label{eq:multi-stage-stage-cost}
\end{equation}
The trajectory cost \(\mathcal J\) is the sum of \(J_t^{\mathrm{exec}}\) over all stages, so each executed period contributes once. Unlike the planning objective, \(\mathcal J\) has no terminal cost. The trajectory covers the whole repositioning window of the day, so \(\mathcal J\) is the realized operating cost of that window.

\subsection{Commitment--Recourse with Demand Forecasts}
\label{sec:commitment-recourse-execution}

The multistage procedure above takes horizon demand as a given input. This subsection replaces it with forecasts. In operation, demand is unknown when repositioning is committed, so the platform relies on forecasts $\widehat d_{ij,t+\ell\mid t}$ for $\ell\in\mathcal H$. Simply substituting these forecasts for demand in the model can produce service decisions that exceed the requests actually available. A forecast-based current-period service plan satisfies \(\mu_{ij,t\mid t}\le\sum_{\tau'\in\mathcal T_t^{\mathrm{arr}}\setminus\{t\}}q_{ij,t,\tau'\mid t}+\widehat d_{ij,t\mid t}\) for each \((i,j)\in\Dpax\), whereas actual execution requires the same bound with the realized demand \(d_{ij,t}\) in place of the forecast. When the forecast exceeds realized demand, the plan can therefore serve more requests than are available. In practice, the platform first commits part of the idle fleet to repositioning and keeps the remaining vehicles available for service. It then chooses ordinary service and eligible pickup conversions under realized demand while retaining that commitment. We model this sequence with two linked LPs. The pre-realization LP uses forecast demand to choose the repositioning commitment. The post-realization LP uses realized current demand to choose service.

In the pre-realization commitment problem, let $\widehat d_t$ stack the nonnegative horizon forecast. Denote the solution map of \eqref{eq:lp-mp} by $\mathrm{LP}_t^{\mathrm{pre}}$. Let $\boldsymbol n_t^{\mathrm{pre}}$ denote the pre reference vector at stage $t$. The fixed selector $\Pi_g$ extracts the current repositioning commitment. The optimal plan and commitment are
\begin{equation}
  u_t^{\mathrm{pre},\star}
  =\mathrm{LP}_t^{\mathrm{pre}}\!\left(s_t,\widehat d_t;
    \boldsymbol n_t^{\mathrm{pre}}\right),
  \qquad g_t^{\mathrm{exec}}=\Pi_g u_t^{\mathrm{pre},\star},
  \qquad \forall t\in\mathcal T.
  \label{eq:pre-lp-solution-map}
\end{equation}

\begin{figure}[H]
  \centering
  \includegraphics[width=0.65\textwidth]{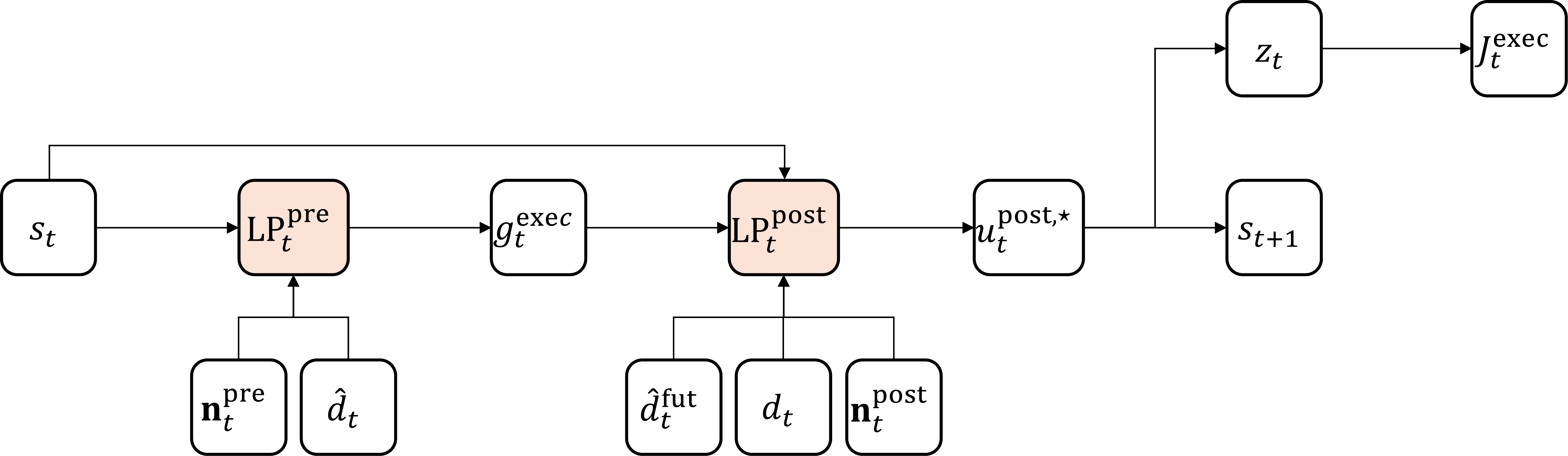}
  \caption{Commitment--recourse policy at stage \(t\). Peach boxes denote the pre- and post-realization LPs, white boxes denote inputs and outputs, and arrows show their dependencies. Both LPs start from the same state \(s_t\). The pre LP uses horizon forecasts \(\widehat d_t\) and the reference vector \(\boldsymbol n_t^{\mathrm{pre}}\) of stocks and accumulations to choose the current repositioning commitment \(g_t^{\mathrm{exec}}\). The post LP retains that commitment, replaces current-period forecast demand with realized demand \(d_t\), and uses the remaining future forecasts \(\widehat d_t^{\mathrm{fut}}\) and post reference vector \(\boldsymbol n_t^{\mathrm{post}}\). Its solution, after cohort reallocation, provides the first-period executed vector \(z_t\), realized cost \(J_t^{\mathrm{exec}}\), and next state \(s_{t+1}\).}
  \label{fig:single-stage-optimization-graph}
\end{figure}

In the post-realization recourse problem, let $d_t=(d_{ij,t})_{(i,j)\in\Dpax}$ collect realized current demand in the same OD order as each forecast block. Once $d_t$ is observed, the post-realization problem uses it at lead zero and retains forecasts at later leads. Let the fixed selector $\Pi_{\mathrm{fut}}$ remove lead zero, so $\widehat d_t^{\mathrm{fut}}=\Pi_{\mathrm{fut}}\widehat d_t$ contains those later forecasts. Solve \eqref{eq:lp-mp} from the same state $s_t$ with demand inputs $(d_t,\widehat d_t^{\mathrm{fut}})$ and the fixed repositioning commitment $g^{\mathrm{post}}_{oj,t\mid t}=g^{\mathrm{exec}}_{oj,t}$. Denote this solution map by $\mathrm{LP}_t^{\mathrm{post}}$. Let $\boldsymbol n_t^{\mathrm{post}}$ denote the corresponding post reference vector. Its optimal solution is
\begin{equation}
  \mathring u_t^{\mathrm{post}}
  =\mathrm{LP}_t^{\mathrm{post}}\!\left(s_t,d_t,\widehat d_t^{\mathrm{fut}},g_t^{\mathrm{exec}};
    \boldsymbol n_t^{\mathrm{post}}\right),
  \qquad \forall t\in\mathcal T.
  \label{eq:post-lp-solution-map}
\end{equation}
This solution provides current ordinary service, pickup, and cohort allocation. The LP decides how much of each arrival cohort to serve. Different splits of the same total service among cohorts can be equally optimal, so the cohort allocation may not be unique. We make it unique with a reallocation map \(\mathcal C_t\). The map keeps the LP's total service and abandonment for each OD pair and period. Within these totals, it serves older cohorts first, after setting aside the requests that abandon at expiry. The map is piecewise affine and does not change the executed cost. Thus the post plan, executed decision, and next state are
\begin{equation}
  u_t^{\mathrm{post},\star}=\mathcal C_t\bigl(\mathring u_t^{\mathrm{post}}\bigr),\quad
  z_t=\Pi_z u_t^{\mathrm{post},\star},\quad
  s_{t+1}=\Pi_s u_t^{\mathrm{post},\star},
  \qquad \forall t\in\mathcal T.
  \label{eq:commitment-recourse-action}
\end{equation}
The trajectory cost \(\mathcal J\) then sums \eqref{eq:multi-stage-stage-cost} over these \(z_t\). This cost includes the repositioning commitment chosen by the pre problem and the service outcomes determined by the post problem. Figure~\ref{fig:single-stage-optimization-graph} summarizes how this sequence links forecast-based repositioning to service under realized demand.

\subsection{Reference-Trajectory Update}
\label{sec:reference-update}
Both optimization layers require reference stocks and accumulations for \eqref{eq:affine-period-event}. A reference whose stocks and accumulations differ substantially from those of the current problem may poorly represent congestion effects. Successive multistage problems have the same model structure and similar demand. Their optimal plans therefore provide useful decision preferences for constructing the references.

We first construct the post reference from the pre optimal plan at the same stage. The two problems differ only in current-period demand, which is forecasted in the pre problem and realized in the post problem. This difference alone prevents the pre optimum from serving directly as the post reference, because it may serve more requests than are available. We therefore keep the repositioning commitment and the pre plan's allocation proportions, and propagate them through the traffic dynamics. A repair step scales service down to the available requests. Appendix~\ref{app:reference-update} gives these steps and the references for the new tail period and the first stage. This construction defines the map
\begin{equation}
  \boldsymbol n_t^{\mathrm{post}}
  =\operatorname{Ref}_t^{\mathrm{post}}(s_t,d_t,\widehat d_t^{\mathrm{fut}},u_t^{\mathrm{pre},\star}),
  \qquad \forall t\in\mathcal T.
  \label{eq:post-reference-map}
\end{equation}

For each stage after initialization, we construct the pre reference from the preceding stage's post optimal plan. The initial state also comes directly from that plan, and the two horizons share $H-1$ periods. 
The plan still cannot be reused directly. Its decisions were made for the previous forecasts and may not fit the updated ones, and it has no decisions for the new final period. We therefore reuse the allocation shares \eqref{eq:reference-idle-allocation} of the preceding post plan in the overlapping periods and carry its final-period shares into the tail. The same repair and propagation then give
\begin{equation}
  \boldsymbol n_t^{\mathrm{pre}}
  =\operatorname{Ref}_t^{\mathrm{pre}}(s_t,\widehat d_t,u_{t-1}^{\mathrm{post},\star}),
  \qquad \forall t\in\mathcal T\setminus\{0\}.
  \label{eq:pre-reference-map}
\end{equation}

\begin{figure}[H]
  \centering
  \includegraphics[width=\textwidth]{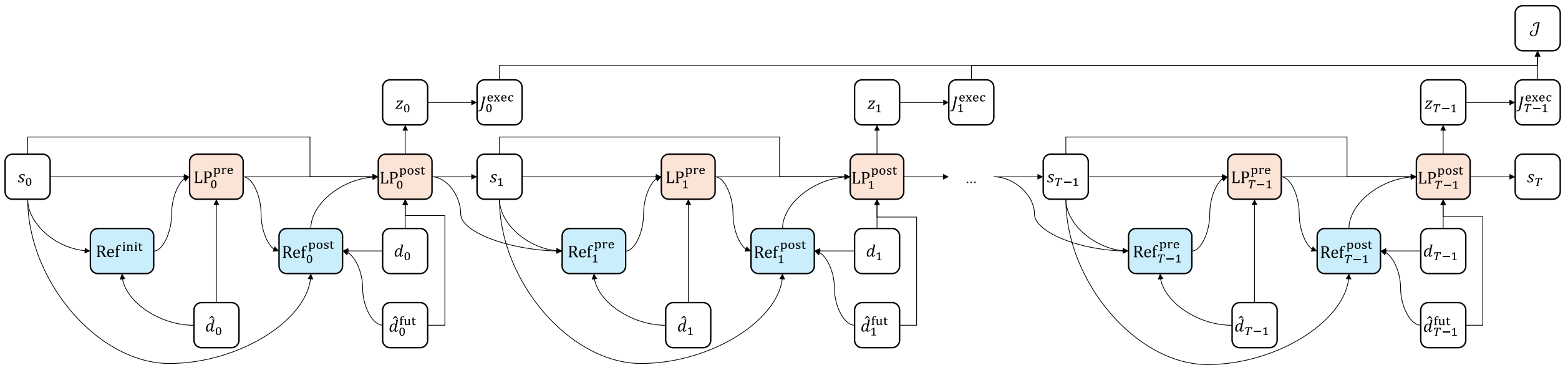}
  \caption{Multistage commitment--recourse execution over \(T\) stages. Peach boxes denote LP solves, blue boxes denote reference-construction maps, and white boxes denote states, demand, decisions, and costs. Black arrows show their input--output dependencies. At stage zero, \(\operatorname{Ref}^{\mathrm{init}}\) constructs the pre reference from \(s_0\) and \(\widehat d_0\). At each stage \(t\), the pre LP uses \(s_t\), the horizon forecast \(\widehat d_t\), and its pre reference to choose the repositioning commitment. The pre optimum also supplies the allocation preferences for \(\operatorname{Ref}_t^{\mathrm{post}}\), which constructs the post reference under realized current demand \(d_t\) and future forecasts \(\widehat d_t^{\mathrm{fut}}\). The post LP retains the commitment and provides the next state \(s_{t+1}\) and executed vector \(z_t\), whose cost \(J_t^{\mathrm{exec}}\) contributes to \(\mathcal J\). Each later pre reference uses the inherited state, updated forecasts, and preceding post plan.}
  \label{fig:multistage-optimization-graph}
\end{figure}

At the first stage, no preceding plan is available. We solve the fixed-fraction LP of Section~\ref{sec:period-level-pwl-mfd-lp} under $s_0$ and $\widehat d_0$. Its optimal plan \(u_0^{\mathrm{init},\star}\) supplies allocation proportions that account for future demand. The same repair and nonlinear propagation give
\begin{equation}
  \boldsymbol n_0^{\mathrm{pre}}
  =\operatorname{Ref}^{\mathrm{init}}(s_0,\widehat d_0,u_0^{\mathrm{init},\star}).
  \label{eq:initial-reference-map}
\end{equation}
Figure~\ref{fig:multistage-optimization-graph} summarizes the multistage policy.

\section{Multistage Contextual Learning and Differentiation}
\label{sec:decision-aware-demand-prediction}
\label{sec:multistage-differentiation}

The previous section defines a multistage commitment--recourse policy. 
We now train a demand predictor for this policy using operating
cost together with prediction error.
We first define the prediction architecture and
computation graph (Section~\ref{sec:prediction-computation-graph}). We then differentiate the LP solutions (Section~\ref{sec:optimal-basis-sensitivity}), combine them with the reference updates and state transitions into stage sensitivities (Section~\ref{sec:single-stage-sensitivity}), and derive equivalent forward and adjoint gradients for training (Section~\ref{sec:multistage-gradients}).
\subsection{Prediction Architecture and Computation Graph}
\label{sec:prediction-computation-graph}

\begin{figure}[htbp]
  \centering
  \includegraphics[width=\textwidth]{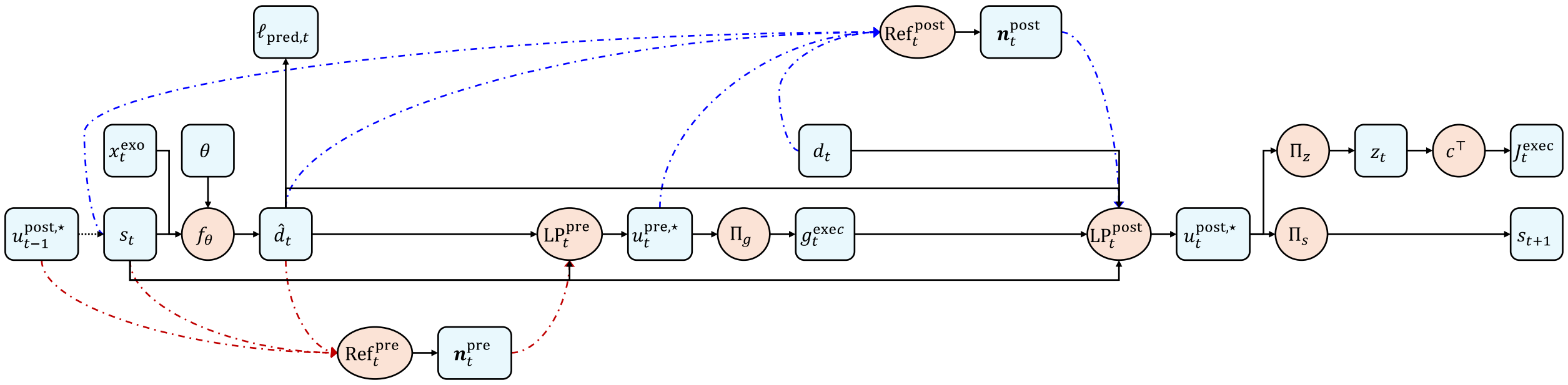}
  \caption{Single-stage computation graph. Blue rectangles denote data, solutions, and loss, peach ellipses denote maps, and black arrows show their input--output dependencies. The predictor \(f_\theta\) maps  \(x_t^{\mathrm{exo}}\) and \(s_t\) to the forecast \(\widehat d_t\), using trainable parameters \(\theta\). Red and blue dash-dotted arrows trace pre- and post-reference construction, respectively. The maps \(\operatorname{Ref}_t^{\mathrm{pre}}\) and \(\operatorname{Ref}_t^{\mathrm{post}}\) produce the reference vectors \(\boldsymbol n_t^{\mathrm{pre}}\) and \(\boldsymbol n_t^{\mathrm{post}}\). The pre solution \(u_t^{\mathrm{pre},\star}\) supplies the repositioning commitment. Given that commitment and realized current demand \(d_t\), the post plan \(u_t^{\mathrm{post},\star}\), obtained from the post solution by cohort reallocation, supplies executed quantities \(z_t\) and the next state. The cost vector \(c\) gives \(J_t^{\mathrm{exec}}=c^\top z_t\). The forecast also determines the prediction loss \(\ell_{\mathrm{pred},t}\). The two loss outputs contribute \(J_t^{\mathrm{exec}}+\alpha\ell_{\mathrm{pred},t}\) to the training objective.}
  \label{fig:single-stage-contextual-graph}
\end{figure}

We treat passenger demand as exogenous to platform operations and use
decision-independent context \(x_t^{\mathrm{exo}}\), such as historical demand, to predict it. However, the operational
effect of a forecast error depends on the state created by earlier decisions.
For example, underpredicting ten requests may have little effect when an
origin has ample idle vehicles, but may suppress needed vehicle inflows and
increase backlog when supply is scarce. 
The predictor therefore uses \(x_t^{\mathrm{exo}}\) together with the
inherited operational state \(s_t\). A differentiable predictor maps both inputs to the demand vector \(\widehat d_t=f_\theta(x_t^{\mathrm{exo}},s_t)\in\mathbb R_+^{H|\Dpax|}\) for \(t\in\mathcal T\), where \(\theta\) is the trainable parameter vector of \(f_\theta\).
Appendix~\ref{app:state-conditioned-predictor} specifies the prediction model.
Let \(\Omega\) be the set of training trajectories, each defined by an
initial state and sequences of realized demand and exogenous context. For
trajectory \(\omega\in\Omega\), let
\(\ell_{\mathrm{pred},t}^\omega(\widehat d_t)\) measure differentiable prediction error at stage \(t\in\mathcal T\). The total prediction
error is \(\mathcal L_{\mathrm{pred}}^\omega(\theta)
=\sum_{t\in\mathcal T}\ell_{\mathrm{pred},t}^\omega(\widehat d_t)\). The
trajectory cost of \(\omega\) is
\(\mathcal J^\omega(\theta)=\sum_{t\in\mathcal T}J_t^{\mathrm{exec},\omega}\). We train the predictor with the objective
\begin{equation}
  \mathcal L(\theta)
  =
  \frac{1}{|\Omega|}
  \sum_{\omega\in\Omega}
  \sum_{t\in\mathcal T}
  \left[
    \underbrace{J_t^{\mathrm{exec},\omega}}_{\text{decision cost}}
    +\alpha
     \underbrace{\ell_{\mathrm{pred},t}^\omega(\widehat d_t)}_{\text{prediction error}}
  \right].
  \label{eq:joint-contextual-training-objective}
\end{equation}
We write \(\mathcal L^\omega(\theta)=\mathcal J^\omega(\theta)+\alpha\mathcal L_{\mathrm{pred}}^\omega(\theta)\) for the loss of sample \(\omega\), so that \(\mathcal L(\theta)\) is its mean over \(\Omega\). The nonnegative weight \(\alpha\) controls the trade-off
between decision cost and prediction error.
The decision cost is the primary
training objective. The prediction error
anchors \(\widehat d_t\) to observed demand.  Figure~\ref{fig:single-stage-contextual-graph} summarizes how forecasts
affect execution through both LP decisions and reference construction,
and contribute to training through the prediction loss.

\subsection{Optimal-Basis Sensitivity}
\label{sec:optimal-basis-sensitivity}

The computation graph combines explicit maps with implicit LP solution
maps. The predictor and selectors can be differentiated directly. The initialization, pre-, and
post-realization solutions require the sensitivity of an optimal LP solution
to its inputs. The state
fixes the initial stocks at each stage. Demand enters the request balances, and the
post-realization LP fixes the current repositioning commitment
\(g_t^{\mathrm{exec}}\). These inputs enter the RHS.
The reference changes the constraint matrix, RHS, and objective coefficients
through the affine movement equations \eqref{eq:affine-period-event}
and the penalty terms in \eqref{eq:penalty-square-epigraph}--\eqref{eq:linear-error-penalty}. Let \(\mathrm{init}\) denote the initialization LP of
Section~\ref{sec:reference-update}, and let
\(\mathcal V_{\mathrm{LP}}=\{(0,\mathrm{init})\}
\cup(\mathcal T\times\{\mathrm{pre},\mathrm{post}\})\) collect the LP nodes.
For each \((t,\sigma)\in\mathcal V_{\mathrm{LP}}\), add nonnegative
slack variables and let \(\widetilde u\) collect the physical and slack
variables.
Let \(\widetilde c_t^\sigma(\eta)\),
\(\widetilde{\mathsf A}_t^\sigma(\eta)\), and \(\widetilde{\mathsf b}_t^\sigma(\eta)\) denote the
equality-form objective coefficients, constraint matrix, and RHS. The vector
\(\eta\) collects the model's inputs listed above.
We write the optimal solution as
\begin{equation}
  \widetilde u_t^{\sigma,\star}(\eta)
  =
  \argmin_{\widetilde u\ge0}
  \left\{
    \widetilde c_t^\sigma(\eta)^\top\widetilde u:
    \widetilde{\mathsf A}_t^\sigma(\eta)\widetilde u
    =
    \widetilde{\mathsf b}_t^\sigma(\eta)
  \right\},
  \qquad \forall (t,\sigma)\in\mathcal V_{\mathrm{LP}}.
  \label{eq:equality-form-parametric-lp}
\end{equation}

\begin{assumption}[Local optimal-basis stability]
\label{ass:local-optimal-basis-stability}
At every LP node in \(\mathcal V_{\mathrm{LP}}\) along the realized trajectory,
there is a neighborhood of the full input vector in which the optimal solution
is selected using the same basis index set and the associated basis matrix remains nonsingular.
\end{assumption}
\begin{assumption}[Fixed branches]
\label{ass:fixed-branches}
Along the realized trajectory, there is a neighborhood of the inputs in which the repair rules in reference construction and the reallocation \(\mathcal C_t\) each stay on the same branch, on which they are smooth.
\end{assumption}

Under Assumptions~\ref{ass:local-optimal-basis-stability} and~\ref{ass:fixed-branches}, every map in the computation graph is locally differentiable, and the gradients derived below are exact. Let
\(\mathcal B_t^\sigma\) and \(\mathcal N_t^\sigma\) be the basic and nonbasic
variable index sets, respectively. Write
\(\mathsf B_t^\sigma(\eta)\)
for the basis matrix. The corresponding primal solution is
\begin{equation}
  \begin{aligned}
  [\widetilde u_t^{\sigma,\star}(\eta)]_{\mathcal B_t^\sigma}
  =
  (\mathsf B_t^\sigma(\eta))^{-1}
  \widetilde{\mathsf b}_t^\sigma(\eta),
  \quad\forall (t,\sigma)\in\mathcal V_{\mathrm{LP}},\\
  [\widetilde u_t^{\sigma,\star}(\eta)]_{\mathcal N_t^\sigma}=0,
  \quad\forall (t,\sigma)\in\mathcal V_{\mathrm{LP}}.
  \end{aligned}
  \label{eq:fixed-basis-primal-selection}
\end{equation}
We write \(\mathsf J^{\mathrm{tag}}_{y,x,t}\) for the Jacobian of output \(y\) with respect to input \(x\) at stage \(t\). The superscript names the LP node (\(\mathrm{init}\), \(\mathrm{pre}\), or \(\mathrm{post}\)) or the derivative path (\(\mathrm{pred}\), \(\mathrm{ref}\), or \(\mathrm{stage}\)) when several Jacobians share an output and an input.
Differentiating the basis equation with respect to \(\eta\) gives
\begin{equation}
  \begin{aligned}
  \mathsf J_{\widetilde u_{\mathcal B},\eta,t}^{\sigma}
  :=
  \frac{\partial
    [\widetilde u_t^{\sigma,\star}]_{\mathcal B_t^\sigma}}
    {\partial\eta}
  ={}
  \underbrace{
    (\mathsf B_t^\sigma)^{-1}
    \frac{\partial \widetilde{\mathsf b}_t^\sigma}{\partial\eta}
  }_{\text{effect of RHS changes}}
  -\underbrace{
    (\mathsf B_t^\sigma)^{-1}
    \frac{\partial\mathsf B_t^\sigma}{\partial\eta}
    [\widetilde u_t^{\sigma,\star}]_{\mathcal B_t^\sigma}
  }_{\text{effect of basis-matrix changes}},
  \forall (t,\sigma)\in\mathcal V_{\mathrm{LP}}.
  \end{aligned}
  \label{eq:fixed-basis-data-sensitivity}
\end{equation}
The derivatives of the nonbasic variables are zero. To return to the physical variables, let
\(\mathsf I_{\mathcal B_t^\sigma}\) place the basic variables into the full
equality-form vector, with zeros elsewhere, and let \(\Pi_u^\sigma\) drop the
slack variables. For any part of \(\eta\), such as the state or the forecast, the Jacobian of the physical solution is
\begin{equation}
  \mathsf J_{u,\eta,t}^\sigma
  :=
  \frac{\partial u_t^{\sigma,\star}}{\partial\eta}
  =
  \Pi_u^\sigma
  \mathsf I_{\mathcal B_t^\sigma}
  \mathsf J_{\widetilde u_{\mathcal B},\eta,t}^{\sigma},
  \qquad\forall (t,\sigma)\in\mathcal V_{\mathrm{LP}},\ \sigma\ne\mathrm{post}.
  \label{eq:fixed-basis-physical-jacobian}
\end{equation}
For the post LP, the same expression gives the Jacobian of the LP solution \(\mathring u_t^{\mathrm{post}}\). The post plan \(u_t^{\mathrm{post},\star}=\mathcal C_t(\mathring u_t^{\mathrm{post}})\) also contains the cohort reallocation of Section~\ref{sec:commitment-recourse-execution}. Let \(\mathsf C_t\) denote the Jacobian of \(\mathcal C_t\) on the fixed branch of Assumption~\ref{ass:fixed-branches}. By the chain rule, the post-plan Jacobian is \(\mathsf C_t\) times the LP-solution Jacobian:
\begin{equation}
  \mathsf J_{u,\eta,t}^{\mathrm{post}}
  =\mathsf C_t\,\Pi_u^{\mathrm{post}}
   \mathsf I_{\mathcal B_t^{\mathrm{post}}}
   \mathsf J_{\widetilde u_{\mathcal B},\eta,t}^{\mathrm{post}},
  \qquad\forall t\in\mathcal T.
  \label{eq:post-reallocation-jacobian}
\end{equation}

\subsection{Single-Stage Sensitivity}
\label{sec:single-stage-sensitivity}

We first differentiate the prediction \(\widehat d_t=f_\theta(x_t^{\mathrm{exo}},s_t)\).
Applying the chain rule gives
\begin{equation}
\begin{aligned}
  \mathsf J_{\widehat d,\theta,t}
  :=
  \frac{\partial\widehat d_t}{\partial\theta}
  &={}
  \underbrace{
    \frac{\partial f_\theta}{\partial x_t^{\mathrm{exo}}}
  }_{\mathsf J_{\widehat d,x^{\mathrm{exo}},t}}
  \underbrace{
    \frac{\partial x_t^{\mathrm{exo}}}{\partial\theta}
  }_{=0}
  +
  \underbrace{
    \frac{\partial f_\theta}{\partial s_t}
  }_{\mathsf J_{\widehat d,s,t}}
  \underbrace{
    \frac{\partial s_t}{\partial\theta}
  }_{\mathsf J_{s,\theta,t}}
  +
  \underbrace{
    \frac{\partial f_\theta}{\partial\theta}
  }_{\mathsf J_{\widehat d,\theta,t}^{\mathrm{pred}}}\\
  &={}
  \underbrace{\mathsf J_{\widehat d,\theta,t}^{\mathrm{pred}}}_{\text{direct parameter effect}}
  +\underbrace{\mathsf J_{\widehat d,s,t}\mathsf J_{s,\theta,t}}
    _{\text{effect through inherited state}},\qquad\forall t\in\mathcal T.
\end{aligned}
  \label{eq:state-conditioned-demand-sensitivity}
\end{equation}

Here \(\mathsf J_{\widehat d,\theta,t}\) is the total forecast derivative. The state Jacobian \(\mathsf J_{s,\theta,t}\) accumulates all earlier
parameter effects through the state trajectory. The initial state is fixed,
so \(\mathsf J_{s,\theta,0}=0\). We next propagate this forecast sensitivity through the commitment--recourse
stage, starting with the pre reference \(\boldsymbol n_t^{\mathrm{pre}}\).
Its inputs are the state \(s_t\), forecast \(\widehat d_t\), and preceding post plan
\(u_{t-1}^{\mathrm{post},\star}\). The preceding post plan depends on \(\theta\) through all earlier stages. Its Jacobian \(\mathsf J_{u,\theta,t-1}^{\mathrm{post}}\) comes from the previous stage of the recursion \eqref{eq:state-conditioned-post-solution-sensitivity} below. Differentiating \eqref{eq:pre-reference-map} gives
\begin{equation}
\begin{aligned}
  \mathsf J_{\boldsymbol n,\theta,t}^{\mathrm{pre}}
  :={}
  \frac{\partial\boldsymbol n_t^{\mathrm{pre}}}{\partial\theta}
  &={}
  \underbrace{\frac{\partial\boldsymbol n_t^{\mathrm{pre}}}{\partial s_t}}
    _{\mathsf J_{\boldsymbol n,s,t}^{\mathrm{pre}}}\mathsf J_{s,\theta,t}
  +\underbrace{\frac{\partial\boldsymbol n_t^{\mathrm{pre}}}{\partial\widehat d_t}}
    _{\mathsf J_{\boldsymbol n,\widehat d,t}^{\mathrm{pre}}}\mathsf J_{\widehat d,\theta,t}
  +\underbrace{\frac{\partial\boldsymbol n_t^{\mathrm{pre}}}{\partial u_{t-1}^{\mathrm{post},\star}}}
    _{\mathsf J_{\boldsymbol n,u,t}^{\mathrm{pre}}}\mathsf J_{u,\theta,t-1}^{\mathrm{post}}\\
  &={}
  \mathsf J_{\boldsymbol n,s,t}^{\mathrm{pre}}\mathsf J_{s,\theta,t}
  +\mathsf J_{\boldsymbol n,\widehat d,t}^{\mathrm{pre}}\mathsf J_{\widehat d,\theta,t}
  +\mathsf J_{\boldsymbol n,u,t}^{\mathrm{pre}}\mathsf J_{u,\theta,t-1}^{\mathrm{post}},
  \qquad\forall t\in\mathcal T\setminus\{0\}.
\end{aligned}
  \label{eq:pre-reference-sensitivity}
\end{equation}
At stage zero, \eqref{eq:initial-reference-map} instead obtains its source
plan \(u_0^{\mathrm{init},\star}\) from the fixed-fraction LP. This plan depends on \(\theta\) only through
\(\widehat d_0\), so its parameter Jacobian is
\(\mathsf J_{u,\theta,0}^{\mathrm{init}}
=\mathsf J_{u,\widehat d,0}^{\mathrm{init}}\mathsf J_{\widehat d,\theta,0}\). Since the initial state is exogenous,
\begin{equation}
  \mathsf J_{\boldsymbol n,\theta,0}^{\mathrm{pre}}
  =\mathsf J_{\boldsymbol n,s,0}^{\mathrm{pre}}
   \underbrace{\mathsf J_{s,\theta,0}}_{=0}
   +\mathsf J_{\boldsymbol n,\widehat d,0}^{\mathrm{pre}}
    \mathsf J_{\widehat d,\theta,0}
   +\mathsf J_{\boldsymbol n,u,0}^{\mathrm{pre}}
    \mathsf J_{u,\theta,0}^{\mathrm{init}}
  =\mathsf J_{\boldsymbol n,\widehat d,0}^{\mathrm{pre}}
    \mathsf J_{\widehat d,\theta,0}
   +\mathsf J_{\boldsymbol n,u,0}^{\mathrm{pre}}
    \mathsf J_{u,\theta,0}^{\mathrm{init}}.
  \label{eq:initial-reference-sensitivity}
\end{equation}

The pre-realization solution map \eqref{eq:pre-lp-solution-map} has three
parameter-dependent inputs:
\(s_t\), \(\widehat d_t\), and its reference. Its gradient and the resulting
commitment gradient from \eqref{eq:pre-lp-solution-map} are
\begin{equation}
\begin{aligned}
  \mathsf J_{u,\theta,t}^{\mathrm{pre}}
  :=
  \frac{\partial u_t^{\mathrm{pre},\star}}{\partial\theta}
  &={}
  \underbrace{
    \frac{\partial u_t^{\mathrm{pre},\star}}{\partial s_t}
  }_{\mathsf J_{u,s,t}^{\mathrm{pre}}}\mathsf J_{s,\theta,t}
  +
  \underbrace{
    \frac{\partial u_t^{\mathrm{pre},\star}}{\partial\widehat d_t}
  }_{\mathsf J_{u,\widehat d,t}^{\mathrm{pre}}}
  \mathsf J_{\widehat d,\theta,t}+
  \underbrace{
    \frac{\partial u_t^{\mathrm{pre},\star}}
    {\partial\boldsymbol n_t^{\mathrm{pre}}}
  }_{\mathsf J_{u,\boldsymbol n,t}^{\mathrm{pre}}}
  \mathsf J_{\boldsymbol n,\theta,t}^{\mathrm{pre}}\\
  &={}
  \mathsf J_{u,s,t}^{\mathrm{pre}}\mathsf J_{s,\theta,t}
  +
  \mathsf J_{u,\widehat d,t}^{\mathrm{pre}}
  \mathsf J_{\widehat d,\theta,t}
  +\mathsf J_{u,\boldsymbol n,t}^{\mathrm{pre}}
   \mathsf J_{\boldsymbol n,\theta,t}^{\mathrm{pre}},\quad\forall t\in\mathcal T,\\
  \mathsf J_{g,\theta,t}
  :=
  \frac{\partial g_t^{\mathrm{exec}}}{\partial\theta}
  &={}
  \Pi_g
  \mathsf J_{u,\theta,t}^{\mathrm{pre}},\qquad\forall t\in\mathcal T.
\end{aligned}
  \label{eq:pre-commitment-sensitivity}
\end{equation}

The post reference in \eqref{eq:post-reference-map} uses \(s_t\), realized demand \(d_t\), future forecasts
\(\widehat d_t^{\mathrm{fut}}\), and the current pre optimum. Its derivative
follows these four input blocks:
\begin{equation}
\begin{aligned}
  \mathsf J_{\boldsymbol n,\theta,t}^{\mathrm{post}}
  &:={}
  \frac{\partial\boldsymbol n_t^{\mathrm{post}}}{\partial\theta}={}
  \underbrace{\frac{\partial\boldsymbol n_t^{\mathrm{post}}}{\partial s_t}}
    _{\mathsf J_{\boldsymbol n,s,t}^{\mathrm{post}}}\mathsf J_{s,\theta,t}
  +\underbrace{\frac{\partial\boldsymbol n_t^{\mathrm{post}}}{\partial d_t}}
    _{\mathsf J_{\boldsymbol n,d,t}^{\mathrm{post}}}
   \underbrace{\frac{\partial d_t}{\partial\theta}}_{=0} +
  \underbrace{\frac{\partial\boldsymbol n_t^{\mathrm{post}}}{\partial\widehat d_t^{\mathrm{fut}}}}
    _{\mathsf J_{\boldsymbol n,\widehat d^{\mathrm{fut}},t}^{\mathrm{post}}}
   \Pi_{\mathrm{fut}}\mathsf J_{\widehat d,\theta,t}
  +\underbrace{\frac{\partial\boldsymbol n_t^{\mathrm{post}}}{\partial u_t^{\mathrm{pre},\star}}}
    _{\mathsf J_{\boldsymbol n,u,t}^{\mathrm{post}}}\mathsf J_{u,\theta,t}^{\mathrm{pre}}\\
  &={}
  \mathsf J_{\boldsymbol n,s,t}^{\mathrm{post}}\mathsf J_{s,\theta,t}
  +\mathsf J_{\boldsymbol n,\widehat d^{\mathrm{fut}},t}^{\mathrm{post}}
   \Pi_{\mathrm{fut}}\mathsf J_{\widehat d,\theta,t}
  +\mathsf J_{\boldsymbol n,u,t}^{\mathrm{post}}\mathsf J_{u,\theta,t}^{\mathrm{pre}},
  \quad\forall t\in\mathcal T.
\end{aligned}
  \label{eq:post-reference-sensitivity}
\end{equation}

The post LP \eqref{eq:post-lp-solution-map} has five inputs: \(s_t\), \(d_t\), \(\widehat d_t^{\mathrm{fut}}\), \(g_t^{\mathrm{exec}}\), and \(\boldsymbol n_t^{\mathrm{post}}\). Realized demand \(d_t\) does not depend on \(\theta\). The commitment and the post reference depend on the pre plan, which depends on the pre reference. Substituting \eqref{eq:pre-commitment-sensitivity}, \eqref{eq:post-reference-sensitivity}, and \eqref{eq:pre-reference-sensitivity} leaves three inputs to the post plan: the forecast, the inherited state, and the preceding post plan. Let \(\mathsf J_{u,\widehat d,t}^{\mathrm{stage}}\) and \(\mathsf J_{u,s,t}^{\mathrm{stage}}\) be the sensitivities to the forecast and the state. The preceding plan acts in two ways. It sets the state and shapes the reference. Together they give \(\mathsf J_{u,u,t}^{\mathrm{stage}}\). Appendix~\ref{app:stage-sensitivity} expands these Jacobians. Substituting the forecast derivative \eqref{eq:state-conditioned-demand-sensitivity} separates the three effects:
\begin{equation}
\begin{aligned}
  \mathsf J_{u,\theta,t}^{\mathrm{post}}
  :={}&\frac{\partial u_t^{\mathrm{post},\star}}{\partial\theta}\\
  ={}&
  \mathsf J_{u,\widehat d,t}^{\mathrm{stage}}\mathsf J_{\widehat d,\theta,t}^{\mathrm{pred}}
  +\underbrace{\left(
    \mathsf J_{u,s,t}^{\mathrm{stage}}\Pi_s
    +\mathsf J_{u,u,t}^{\mathrm{ref}}
  \right)}_{\mathsf J_{u,u,t}^{\mathrm{stage}}}
  \mathsf J_{u,\theta,t-1}^{\mathrm{post}}
  +\mathsf J_{u,\widehat d,t}^{\mathrm{stage}}\mathsf J_{\widehat d,s,t}\mathsf J_{s,\theta,t}\\
  ={}&
  \underbrace{\mathsf J_{u,\widehat d,t}^{\mathrm{stage}}\mathsf J_{\widehat d,\theta,t}^{\mathrm{pred}}}
    _{\substack{\text{parameters change}\\\text{prediction and decision}}}
  +\underbrace{\mathsf J_{u,u,t}^{\mathrm{stage}}\mathsf J_{u,\theta,t-1}^{\mathrm{post}}}
    _{\substack{\text{previous plan changes state,}\\\text{reference, and decision}}}
  +\underbrace{\mathsf J_{u,\widehat d,t}^{\mathrm{stage}}\mathsf J_{\widehat d,s,t}\mathsf J_{s,\theta,t}}
    _{\substack{\text{state changes prediction}\\\text{and decision}}},
  \forall t\in\mathcal T\setminus\{0\}.
\end{aligned}
  \label{eq:state-conditioned-post-solution-sensitivity}
\end{equation}

The first term follows the prediction-to-decision path of classical contextual optimization \citep{ElmachtoubGrigas2022}: at fixed context,
parameter changes alter the forecast and hence the decision. 
The second term is added by the multistage setting. It couples the stages: changes in the preceding optimal plan affect the current decision through the inherited state and reference construction.
The third term is added by decision-dependent context. The inherited state depends on earlier decisions and also serves as prediction context, so changes in this state alter the forecast and then the current decision.

At stage zero, the state-gradient contributions vanish because
\(\mathsf J_{s,\theta,0}=0\). Here,
\(\mathsf J_{u,u,0}^{\mathrm{ref}}\) denotes only the effect of the
initialization plan through reference construction, and acts on
\(\mathsf J_{u,\theta,0}^{\mathrm{init}}\). This leaves two paths:
the forecast influences the decision directly through the stage map, and it
alters the initialization plan, which shapes the reference and hence the
decision:
\begin{equation}
  \mathsf J_{u,\theta,0}^{\mathrm{post}}
  =\underbrace{
    \mathsf J_{u,\widehat d,0}^{\mathrm{stage}}
    \mathsf J_{\widehat d,\theta,0}^{\mathrm{pred}}
  }_{\text{direct forecast path}}
  +\underbrace{
    \mathsf J_{u,u,0}^{\mathrm{ref}}
    \mathsf J_{u,\theta,0}^{\mathrm{init}}
  }_{\text{initialization-plan path}}.
  \label{eq:initial-post-solution-sensitivity}
\end{equation}

\subsection{Multistage Forward and Adjoint Gradients}
\label{sec:multistage-gradients}

The fixed selectors \(\Pi_z\) and \(\Pi_s\) map the post-realization solution
to the executed vector and next state in \eqref{eq:commitment-recourse-action}.
The next stage also reuses the post plan through \eqref{eq:pre-reference-map}.
Differentiating the selectors gives the forward recursion
\begin{equation}
  \begin{aligned}
  \mathsf J_{z,\theta,t}
  :=
  \frac{\partial z_t}{\partial\theta}
  =
  \Pi_z
  \mathsf J_{u,\theta,t}^{\mathrm{post}},
  &\qquad
  \mathsf J_{s,\theta,t+1}
  =
  \Pi_s
  \mathsf J_{u,\theta,t}^{\mathrm{post}},
  &\qquad\forall t\in\mathcal T.
  \end{aligned}
  \label{eq:full-state-sensitivity-recursion}
\end{equation}
For trajectory \(\omega\in\Omega\), combining the executed-cost
and prediction-loss branches of \eqref{eq:joint-contextual-training-objective} gives the forward gradient
\begin{equation}
\begin{aligned}
  \nabla_\theta\mathcal L^\omega(\theta)
  =
  \sum_{t\in\mathcal T}
  \left[
    \left(
      \mathsf J_{z,\theta,t}
    \right)^\top c
    +
    \alpha
    \left(
      \mathsf J_{\widehat d,\theta,t}
    \right)^\top
    \nabla_{\widehat d_t}
    \ell_{\mathrm{pred},t}^\omega
  \right],\qquad\forall\omega\in\Omega.
\end{aligned}
  \label{eq:full-forward-pathwise-gradient}
\end{equation}

\begin{figure}[htbp]
  \centering
  \includegraphics[width=0.65\textwidth]{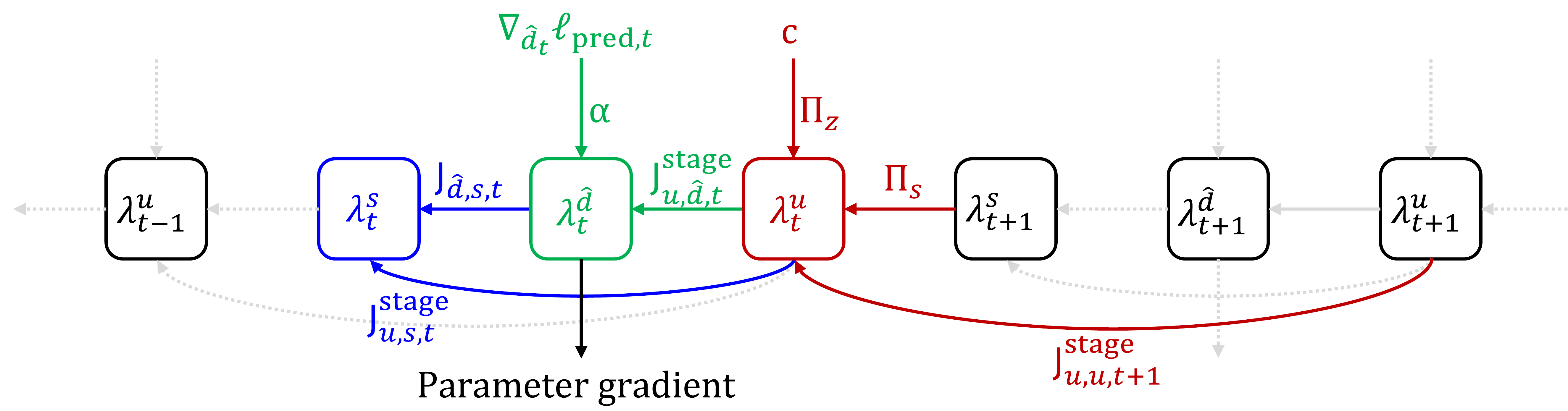}
  \caption{Adjoint recursion at an interior stage of the multistage policy. Rounded rectangles contain the accumulated loss gradients with respect to the post plan, forecast, and state, denoted by \(\lambda_t^u\), \(\lambda_t^{\widehat d}\), and \(\lambda_t^s\). Red paths combine the current cost gradient \(c\) with future contributions through the next state and the retained plan used for reference updating. Green paths combine the decision contribution with the prediction-error gradient weighted by \(\alpha\). Blue paths return the decision and prediction contributions to the inherited state. Gray paths show the same recursion at adjacent stages. Arrow labels denote forward Jacobians or selection matrices; backward propagation uses their transposes. The black downward arrow contributes \((\mathsf J_{\widehat d,\theta,t}^{\mathrm{pred}})^\top\lambda_t^{\widehat d}\) to the parameter gradient. Summing these contributions over stages gives the trajectory gradient.}
  \label{fig:adjoint-recursion}
\end{figure}

The forward
recursion propagates the total state--parameter
Jacobian \(\mathsf J_{s,\theta,t}\) and retains the preceding post-solution
Jacobian at every stage. Because training needs only the gradient of a scalar
objective, forming these full matrices is unnecessary. 
We therefore derive an adjoint recursion that computes the training gradient
by propagating vector--Jacobian products backward through the computation graph.
Let \(\lambda_t^u\), \(\lambda_t^{\widehat d}\), and \(\lambda_t^s\)
denote the accumulated current-and-future loss gradients with respect to
\(u_t^{\mathrm{post},\star}\), \(\widehat d_t\), and \(s_t\), respectively.
No later stage uses \(s_T\) or \(u_{T-1}^{\mathrm{post},\star}\), so
\(\lambda_{T-1}^u=\Pi_z^\top c\). At earlier stages, the post
plan affects the current executed cost through \(z_t\) and future losses
through both \(s_{t+1}\) and the next reference's preceding-plan input.
The plan adjoint is therefore
\begin{equation}
  \lambda_t^u
  =
  \underbrace{\Pi_z^\top c}_{\text{current executed cost}}
  +
  \underbrace{
    \Pi_s^\top\lambda_{t+1}^s
  }_{\text{future loss through }s_{t+1}}
  +\underbrace{
    (\mathsf J_{u,u,t+1}^{\mathrm{ref}})^\top
    \lambda_{t+1}^u
  }_{\text{future loss through reference update}},
  \quad\forall t\in\mathcal T\setminus\{T-1\}.
  \label{eq:solution-node-adjoint}
\end{equation}
The predictor output affects the loss through the post plan
and the prediction-error term. Its adjoint is
\begin{equation}
  \lambda_t^{\widehat d}
  =
  \underbrace{
    (\mathsf J_{u,\widehat d,t}^{\mathrm{stage}})^\top\lambda_t^u
  }_{\text{decision branch}}
  +
  \underbrace{
    \alpha
    \nabla_{\widehat d_t}\ell_{\mathrm{pred},t}
  }_{\text{prediction-error branch}},\qquad\forall t\in\mathcal T\setminus\{0\}.
  \label{eq:predictor-output-adjoint}
\end{equation}
At stage zero, the forecast also changes the initialization plan, so
\begin{equation}
  \lambda_0^{\widehat d}
  =\left(\mathsf J_{u,\widehat d,0}^{\mathrm{stage}}
    +\mathsf J_{u,u,0}^{\mathrm{ref}}
     \mathsf J_{u,\widehat d,0}^{\mathrm{init}}\right)^\top\lambda_0^u
   +\alpha\nabla_{\widehat d_0}\ell_{\mathrm{pred},0}.
  \label{eq:initial-predictor-output-adjoint}
\end{equation}
The inherited state affects the same-stage decisions and changes the predictor output. These two paths give the backward recursion
\begin{equation}
  \lambda_t^s
  =
  \underbrace{
    (\mathsf J_{u,s,t}^{\mathrm{stage}})^\top\lambda_t^u
  }_{\text{direct decision path}}
  +
  \underbrace{
    \mathsf J_{\widehat d,s,t}^\top\lambda_t^{\widehat d}
  }_{\text{state-conditioned prediction path}},
  \qquad
  \forall t\in\mathcal T\setminus\{0\},
  \label{eq:multistage-adjoint-recursion}
\end{equation}
evaluated in reverse stage order. Summing the direct predictor--parameter
contributions gives the trajectory gradient
\(\sum_{t\in\mathcal T}
(\mathsf J_{\widehat d,\theta,t}^{\mathrm{pred}})^\top\lambda_t^{\widehat d}\).
The sample-mean gradient \(\nabla_\theta\mathcal L(\theta)\) is the average
of this expression over \(\Omega\).
Figure~\ref{fig:adjoint-recursion} summarizes how future losses propagate
through the inherited state and reference updates and combine with current
decision cost and prediction error. It also shows how each stage contributes
to the parameter gradient.

\section{Parallel Block Evaluation for Multistage Learning}
\label{sec:parallel-trajectory}

The forward trajectory requires two LP solves per stage, in sequence, and training repeats it after every parameter update. These sequential solves take a large part of training time. We present parallel block evaluation, which accelerates the forward evaluation of the multistage policy. Nearby parameter values produce similar trajectories, so an earlier evaluation of the same sample gives approximate inputs for all stages. We first write the policy as a stage recursion and group contiguous stages into blocks (Section~\ref{sec:trajectory-consistency}). The blocks are then evaluated in parallel, and repeatedly passing each block's output to the next block restores the links between blocks (Section~\ref{sec:parallel-correction}). An adjoint through all stages gives the training direction (Section~\ref{sec:inexact-trajectory-training}). Section~\ref{sec:trajectory-guarantees} gives the properties of the algorithm.

\subsection{Stage Recursion and Block Maps}
\label{sec:trajectory-consistency}

Successive stages depend on one another only through the preceding post optimal plan. This plan determines the next state and supplies the source plan for the next pre reference. We write \(u_t=u_t^{\mathrm{post},\star}\) for \(t\in\mathcal T\). For fixed sample data, let \(F_t\) denote the prediction, reference construction, and commitment--recourse optimization at stage \(t\in\mathcal T\). The first stage uses the given state \(s_0\) and the initialization LP, so \(u_0=F_0(\theta)\). Each later stage uses the preceding post plan, so \(u_t=F_t(u_{t-1};\theta)\) for \(t\in\mathcal T\setminus\{0\}\). To expose parallel work, we partition the stages into \(M\) contiguous blocks with boundary stages \(0=\bar t_0<\bar t_1<\cdots<\bar t_M=T\), where \(1\le M\le T\). Block \(j\in\mathcal M:=\{0,\ldots,M-1\}\) contains the stages \(\mathcal S_j=\{\bar t_j,\ldots,\bar t_{j+1}-1\}\). The output of block \(j\) carries the index \(j+1\) of the next block. Each block evaluates its stages exactly as serial evaluation does. Its block map \(\mathcal F_j=F_{\bar t_{j+1}-1}\circ\cdots\circ F_{\bar t_j}\) composes these stage maps in their original order and returns the last plan of the block.

\subsection{Parallel Block Evaluation}
\label{sec:parallel-correction}

\begin{figure}[htbp]
  \centering
  \includegraphics[width=\textwidth]{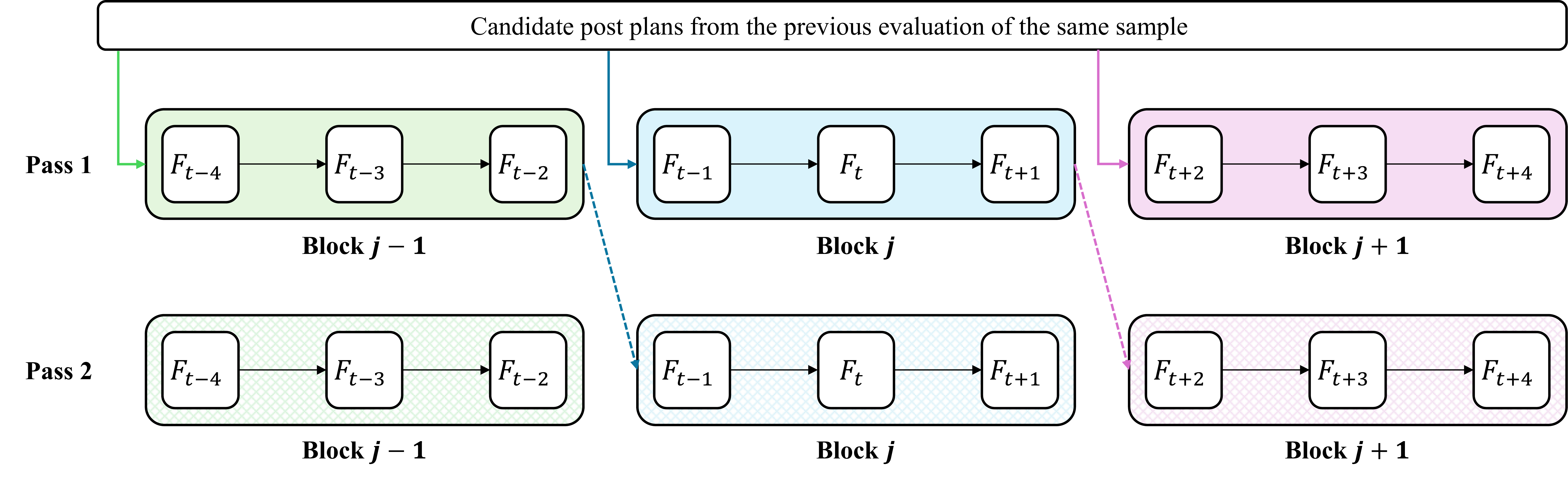}
  \caption{Parallel block evaluation for three consecutive blocks over two passes. Each inner rounded box \(F_t\) is a stage map. Black arrows inside a block pass each post plan to the next stage. Fill colors identify blocks. Each colored arrow has the color of the block it enters. Solid arrows carry candidate post plans from the previous evaluation of the same sample to the block entries in pass~1. Dashed arrows carry the last plan of each block in pass~1 to the entry of the next block in pass~2. Blocks in the same pass are solved in parallel.}
  \label{fig:parallel-algorithm}
\end{figure}

We now solve the blocks in parallel. Under serial evaluation, block \(j\) must wait for the last plan of block \(j-1\). The parameters change gradually during training. The trajectory of a sample therefore tends to stay close to the trajectory from its previous evaluation. That evaluation gives each block an approximate entry plan before the true one is known. All blocks can then be solved at the same time, as shown in Figure~\ref{fig:parallel-algorithm}.

Block \(0\) starts from the given state \(s_0\). For every other block \(j\in\mathcal M\setminus\{0\}\), let \(\bar u_j\) be its entry plan. At first, \(\bar u_j\) is the candidate from the previous evaluation. Each block solves its stages in order. It returns its last plan \(\widehat u_{j+1}\):
\begin{equation}
  \widehat u_1=\mathcal F_0(\theta),
  \qquad
  \widehat u_{j+1}=\mathcal F_j(\bar u_j;\theta),
  \qquad\forall j\in\mathcal M\setminus\{0\}.
  \label{eq:trajectory-stage-output}
\end{equation}
The candidates contain errors. The output \(\widehat u_{j+1}\) generally differs from the entry \(\bar u_{j+1}\). We therefore repeat the solve in passes. In each pass, the output of each block becomes the entry of the next block in the following pass (pass~2 in Figure~\ref{fig:parallel-algorithm}). Let \(\kappa\in\mathbb N=\{1,2,\ldots\}\) index the passes. The parameters \(\theta\) stay fixed. We add the superscript \((\kappa)\) to the entries and outputs of pass \(\kappa\). Each pass applies \eqref{eq:trajectory-stage-output} to the blocks it solves. The entries are updated by
\begin{equation}
  \bar u_{j+1}^{(\kappa+1)}=\widehat u_{j+1}^{(\kappa)},
  \qquad\forall j\in\mathcal M\setminus\{M-1\},\ \kappa\in\mathbb N.
  \label{eq:trajectory-round-transition}
\end{equation}

A block needs no new solve once its entry stops changing. Block \(0\) is solved only in pass \(1\). Block \(1\) has the same entry from pass \(2\) on. In general, block \(j\) has the same entry from pass \(j+1\) on. We call block \(j\) confirmed after its solve in pass \(j+1\). Pass \(\kappa\) thus solves only the blocks in \(\mathcal M_\kappa=\{j\in\mathcal M:j\ge\kappa-1\}\). For a block \(j\notin\mathcal M_\kappa\), the output carries over from the previous pass: \(\widehat u_{j+1}^{(\kappa)}=\widehat u_{j+1}^{(\kappa-1)}\). Let \(K\) denote the number of passes performed. The final records are the stage inputs, LP solutions, and optimal bases from the last solve of each block. Stages inside a block connect by construction. Mismatches appear only at block boundaries. The boundary residual after block \(j\in\mathcal M\setminus\{M-1\}\) in pass \(\kappa\) is \(r_j^{(\kappa)}=\bar u_{j+1}^{(\kappa)}-\widehat u_{j+1}^{(\kappa)}\). The stopping rule can end the passes after a fixed number of passes. It can also end them once the residuals of selected physical quantities fall below a preset threshold.

\subsection{Training Direction from an Evaluated Trajectory}
\label{sec:inexact-trajectory-training}

\begin{figure}[htbp]
  \centering
  \includegraphics[width=\textwidth]{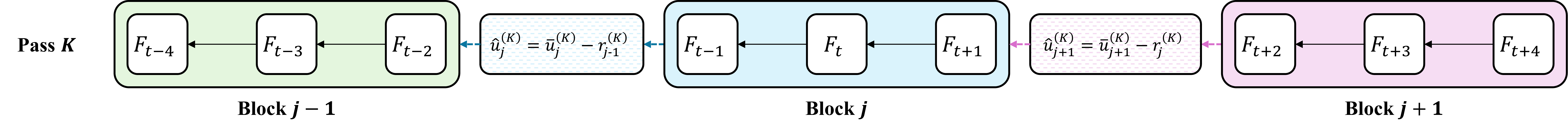}
  \caption{Backward pass on the final records after pass \(K\) for three consecutive blocks. Each inner rounded box \(F_t\) is a stage map. Black arrows inside a block carry the adjoint from each stage to the preceding stage. A hatched box between two blocks states that the output of the left block equals the entry of the right block minus the fixed residual. Dashed arrows carry the adjoint from the first stage of the right block through this relation to the last stage of the left block. Fill colors identify blocks. Each hatched box and dashed arrow has the color of the block on its right.}
  \label{fig:parallel-backward}
\end{figure}

The backward pass runs on the final records of pass \(K\). Inside a block, each stage reads the output of the preceding stage. At the boundary after block \(j\), block \(j+1\) reads its entry \(\bar u_{j+1}^{(K)}\). This entry differs from the block output \(\widehat u_{j+1}^{(K)}\) by the residual \(r_j^{(K)}\). We hold these residuals fixed. Each entry \(\bar u_{j+1}^{(K)}\) then equals the output \(\widehat u_{j+1}^{(K)}\) plus a constant. Its derivative with respect to that output is the identity. The adjoint recursion \eqref{eq:solution-node-adjoint}--\eqref{eq:multistage-adjoint-recursion} therefore applies without change at the final records (Figure~\ref{fig:parallel-backward}). Its parameter contributions give the training direction \(\widetilde\nabla\). When all residuals vanish, \(\widetilde\nabla\) is the exact gradient. Otherwise, \(\widetilde\nabla\) is the gradient of the training loss with these residuals held fixed.

Algorithm~\ref{alg:parallel-trajectory} summarizes the forward passes and the backward pass for one training sample.

\begin{algorithm}[htbp]
\caption{Parallel block evaluation and training direction for one sample.}
\label{alg:parallel-trajectory}
\begin{minipage}{\linewidth}
\textbf{Input:} Fixed \(\theta\) and sample data; boundary stages \(\bar t_0,\ldots,\bar t_M\); candidate entries \(\bar u_j^{(1)}\) for \(j\in\mathcal M\setminus\{0\}\); stopping rule. Set \(\kappa=1\).
\begin{enumerate}[leftmargin=1.8em,itemsep=0.35em,topsep=0.4em]
  \item Solve the blocks in \(\mathcal M_\kappa\) in parallel by \eqref{eq:trajectory-stage-output} using \(\bar u_j^{(\kappa)}\).
  \item Compute the boundary residuals \(r_j^{(\kappa)}\).
  \item If \(\kappa=M\) or the stopping rule ends the passes, set \(K=\kappa\). Go to step 4. Otherwise, form the next entries by \eqref{eq:trajectory-round-transition}. Increase \(\kappa\) by one. Return to step 1.
  \item Apply the adjoint recursion of Section~\ref{sec:decision-aware-demand-prediction} to the final records. Return the training direction \(\widetilde\nabla\).
\end{enumerate}
\textbf{Output:} Final records, boundary residuals, and the training direction \(\widetilde\nabla\).
\end{minipage}
\end{algorithm}
\FloatBarrier

\subsection{Algorithm Properties}
\label{sec:trajectory-guarantees}

Parallel block evaluation gains speed by stopping after a few passes, so its trajectory and training direction are approximate. Three questions then decide whether it can replace serial evaluation in training. First, do the passes reproduce the serial trajectory if they run long enough? If so, the algorithm approximates the serial policy rather than a different one. Second, how do the errors at the block entries change from pass to pass? The answer guides the choice of the number of blocks and passes. Third, how do these forward errors affect the training direction, which the parameter update uses? Proposition~\ref{prop:trajectory-recovery} answers the three questions. Let \((u_t^{\mathrm{ser}})_{t\in\mathcal T}\) denote the serial plans given by the stage recursion of Section~\ref{sec:trajectory-consistency}.

\begin{proposition}[Properties of parallel block evaluation]
\label{prop:trajectory-recovery}
\leavevmode
\begin{enumerate}[label=(\roman*),leftmargin=2em,itemsep=0.2em,topsep=0.3em]
  \item \emph{Finite recovery.} Suppose each stage map returns the same plan for the same inputs, independently of solver warm starts. Then at most \(M\) passes reproduce the serial trajectory exactly.
  \item \emph{Error propagation.} For \(t\in\mathcal T\setminus\{0\}\), let \(L_t\) be a Lipschitz constant of \(F_t\). In pass \(\kappa\), the entry error of block \(j\) is bounded by a multiple of the candidate error \(\kappa-1\) blocks upstream:
  \begin{equation}
    \bigl\|\bar u_j^{(\kappa)}-u_{\bar t_j-1}^{\mathrm{ser}}\bigr\|
    \le\Bigl(\prod_{t=\bar t_{j-\kappa+1}}^{\bar t_j-1}L_t\Bigr)\,
    \bigl\|\bar u_{j-\kappa+1}^{(1)}-u_{\bar t_{j-\kappa+1}-1}^{\mathrm{ser}}\bigr\|,
    \qquad\forall\kappa\in\{1,\ldots,K\},\ j\in\{\kappa,\ldots,M-1\}.
    \label{eq:trajectory-pass-error}
  \end{equation}
  \item \emph{Gradient error.} Suppose Assumptions~\ref{ass:local-optimal-basis-stability} and~\ref{ass:fixed-branches} hold between the final records and serial evaluation, and the stage maps and stage losses have bounded and Lipschitz first derivatives near the serial trajectory. Then the parallel training direction \(\widetilde\nabla\) and the exact serial gradient \eqref{eq:full-forward-pathwise-gradient} satisfy
  \begin{equation}
    \bigl\|\widetilde\nabla-\nabla_\theta\mathcal L^\omega(\theta)\bigr\|
    \le\Lambda\max_{j\in\mathcal M\setminus\{0\}}\bigl\|\bar u_j^{(K)}-u_{\bar t_j-1}^{\mathrm{ser}}\bigr\|,
    \qquad\forall\omega\in\Omega.
    \label{eq:trajectory-gradient-error}
  \end{equation}
\end{enumerate}
\end{proposition}

Appendix~\ref{app:parallel-trajectory-proof} proves these statements. Part~(i) answers the first question. Each pass confirms at least one more block, so the passes reproduce the serial trajectory for any partition into blocks. Recovery alone gives no speedup, however. 
The speedup comes from stopping after a few passes. Part~(ii) answers the second question. Intuitively, an error in a candidate entry travels downstream by one block per pass, and each stage it crosses scales it by at most its Lipschitz constant \(L_t\). When the product of the \(L_t\) over each block is below one, every pass shrinks the error that reaches a block, and a few passes suffice. When the product exceeds one, the error can grow, and more passes or shorter blocks are needed. Part~(iii) answers the third question. The passes control only forward quantities, namely the plans at the block entries. Part~(iii) shows that this suffices: the error of the training direction is at most a constant times the largest entry error in the last pass. Accurate entries therefore give an accurate training direction.

\section{Experiments}
\label{sec:experiments}

\subsection{Experimental Setup}
\label{sec:experimental-setup}

\begin{figure}[htbp]
  \centering
  \includegraphics[width=\textwidth]{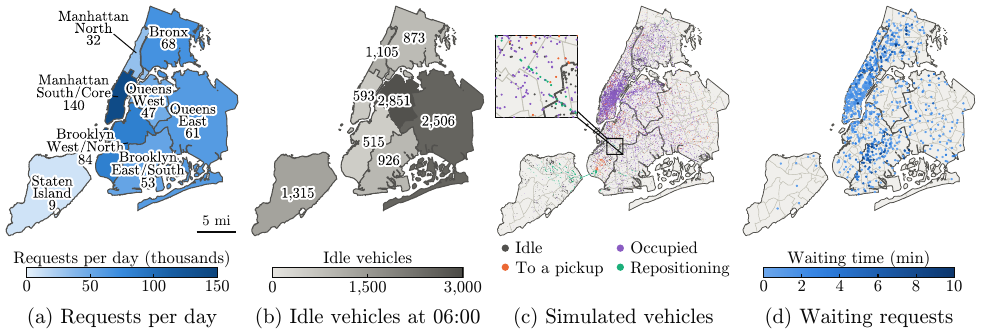}
  \caption{Study area. (a) The eight regions of the planning model, shaded by their mean requests per day (thousands) on the 20 test days, 06:00--21:00. (b) Mean idle vehicles at 06:00 by region on the test days. (c) The taxi zones of simulation and every vehicle at 08:00 on April 7, just before that minute's matching, colored by its status. (d) The requests waiting to be matched at the same moment, at their pickup points and shaded by the time they have waited.}
  \label{fig:study-area}
\end{figure}

The experiments use New York City ride-hailing trip data \citep{NYCTLC2026}. The city is partitioned into eight regions (Figure~\ref{fig:study-area}a), which gives 64 regional OD pairs. Requests are aggregated into five-minute counts. 
Appendix~\ref{app:data} gives further details on the data and the model parameters. 
Repositioning is planned from 06:00 to 21:00 each day, which gives 180 decision stages. The data cover January 1 to April 30, 2026. January and February serve only as history for the forecasts. March (31 days) is used for training, April 1--10 for validation, and April 11--30 (20 days) for testing. All experiments share the same fleet, MFD curves, routes, background traffic, and daily initial states.

The multistage optimization uses five-minute periods, a 30-minute horizon (\(H=6\)), and a waiting allowance \(K^{\mathrm{wait}}=2\). The horizon covers a typical trip: passenger trips in March last 19 minutes on average. Most vehicles dispatched at a stage therefore finish their trips within the horizon, while the stage LPs stay small enough to be solved and differentiated many times in training. With the waiting allowance, a request can be served until the start of the second period after its arrival. Its patience is therefore 7.5 minutes on average. Costs and rewards are measured in US dollars. Waiting costs \$0.67 per request-minute and repositioning costs \$0.56 per committed mile \citep{USDOT2026BCA}. We set the abandonment cost to \$6.70, equivalent to ten minutes of waiting. Completion rewards use the January--February mean fare minus driver pay for each OD pair. The planning objective also contains the movement-error penalty, with weight \(c^{\mathrm{mfd}}=1{,}000\), and the terminal cost, with \(\nu=0.01\). LPs are solved with HiGHS 1.12.0 using dual simplex.

We evaluate all methods in an agent-based simulation of individual vehicles and requests. Every five minutes, a method decides from the aggregate state how many requests to serve, how many idle vehicles to reposition, and how many repositioning vehicles pick up en route. The simulation assigns these numbers to individual vehicles and recorded requests in the taxi zones of the city (Figures~\ref{fig:study-area}c and~\ref{fig:study-area}d), moves the vehicles at the regional MFD speeds, and returns the resulting state to the method at the next stage. The multistage performance of each method is therefore measured on the state of the simulated system rather than that of its planning model. Appendix~\ref{app:simulation} describes the simulation. Validation, which selects the hyperparameters and checkpoints of every method, uses five-minute execution with one nonlinear MFD update per period. All test results in the main text come from the simulation.

\subsection{MFD Optimization Model}
\label{sec:experimental-results}

We first examine the operational model itself. We compare the planning LP \eqref{eq:lp-mp} with the fixed-fraction LP of Section~\ref{sec:period-level-pwl-mfd-lp} and measure the value of repositioning and pickup. These experiments use two demand forecasts. The historical-average (HA) forecast is the January--February mean by weekday, five-minute clock bin, and OD pair. The oracle uses the realized demand and removes forecast error. We report four operating metrics. Business benefit is the completion reward minus the waiting, abandonment, and repositioning costs, per request. Waiting time is the mean time from a request's arrival to its match, its abandonment, or the end of the day. Abandonment and completions are percentages of requests. Table~\ref{tab:planner-ablation} reports the results on the 20 test days.

The planning LP outperforms the fixed-fraction LP with both forecasts. A more accurate forecast helps both planners: the oracle raises the benefit by 3.57\% with the fixed-fraction LP and by 3.48\% with the planning LP. Repositioning and en-route pickup add more. Without repositioning, the benefit falls by 12.16\% with HA and by 14.29\% with the oracle. Without pickup, it falls by 6.44\% and 7.00\%. This loss exceeds the direct value of the pickups: even if every pickup saved a request from abandonment, the pickups would explain at most a third of it. About half of the repositioning trips carry a passenger en route, so without pickups every repositioning trip runs empty, and the planning LP repositions about half as much (17,796 against 38,477 miles per day with HA). The reduced repositioning causes most of the loss through more waiting and abandonment.

\begin{table}[htbp]
  \centering
  \caption{Closed-Loop Performance of the Planning LP and Its Variants}
  \label{tab:planner-ablation}
  \small
  \setlength{\tabcolsep}{4pt}
  \begin{tabular}{lrrrr}
    \toprule
    Variant & \shortstack{Benefit (USD)} & \shortstack{Waiting time (min)} & \shortstack{Abandonment (\%)} & \shortstack{Completions (\%)} \\
    \midrule
    \multicolumn{5}{l}{\emph{HA}} \\
    Planning LP & 2.554 & 1.54 & 12.09 & 86.41 \\
    Fixed-fraction LP & 2.522 $(-1.27\%)$ & 1.60 & 12.22 & 86.27 \\
    No repositioning & 2.244 $(-12.16\%)$ & 1.66 & 14.26 & 84.26 \\
    No pickup & 2.390 $(-6.44\%)$ & 1.60 & 13.20 & 85.31 \\
    \midrule
    \multicolumn{5}{l}{\emph{Oracle}} \\
    Planning LP & 2.643 & 1.49 & 11.62 & 86.89 \\
    Fixed-fraction LP & 2.612 $(-1.18\%)$ & 1.54 & 11.78 & 86.72 \\
    No repositioning & 2.266 $(-14.29\%)$ & 1.64 & 14.24 & 84.29 \\
    No pickup & 2.458 $(-7.00\%)$ & 1.56 & 12.87 & 85.65 \\
    \bottomrule
  \end{tabular}
  \par\smallskip
  \begin{minipage}{\linewidth}\footnotesize\emph{Note.} 
  	Parentheses give the relative change in benefit over the planning LP with the same forecast, computed from unrounded values.\end{minipage}
\end{table}

\subsection{Multistage Contextual Optimization}
\label{sec:learning-results}

We next compare multistage contextual optimization (MCO) with the baselines in Table~\ref{tab:learning-sim}. All methods share the execution, the data splits, and the metrics. Their hyperparameters are tuned on the validation days and fixed before testing. Forecast root-mean-square error (RMSE) is measured in requests per OD pair and five-minute period, pooled over all horizon leads, stages, and days. We consider baselines from five paradigms, grouped as in the table. Appendix~\ref{app:experiment-details} gives their implementation and training details.

Predict-then-optimize methods pass forecasts trained on prediction error to the same policy. They are MSE, our predictor (Appendix~\ref{app:state-conditioned-predictor}) trained by mean squared error without state inputs; HA; STID, the spatial--temporal forecaster \citep{ShaoEtAl2022STID}; ST-SSDL, which learns how demand deviates from its history \citep{GaoEtAl2025STSSDL}; MAGE, which combines learned graphs of the OD pairs \citep{MaEtAl2025MAGE}; and the oracle. Optimization under uncertainty replaces the forecast by historical demand scenarios \citep{BertsimasKallus2020}. SAA, CVaR, and RO minimize the mean, the conditional value-at-risk \citep{RockafellarUryasev2000}, and the worst case of the scenario costs, respectively. Heuristic rules use no optimization model. They give priority to service: idle vehicles first serve waiting requests, and a rule then repositions the remaining idle vehicles. Forecast-free rules use only the current state and recent requests: the +1 rule \citep{RuchEtAl2020PlusOne}, equal distribution \citep{GammelliEtAl2021}, least-squares rebalancing \citep{ValadkhaniRamezani2023}, random walk \citep{GuoEtAl2023}, lost-request dispatch \citep{BeojoneEtAl2024}, and real-time rebalancing \citep{PavoneEtAl2012}. Forecast-based rules set targets from a forecast: forecast-guided greedy dispatch with MSE forecasts \citep{GuoEtAl2026}; historical-share targets from HA; and target cruise toward the region with the highest MSE forecast \citep{GuoEtAl2023}. Among reinforcement learning (RL) methods, learned-value lookahead adds to the planning LP of MSE a value for each vehicle's end region at the horizon, learned by temporal-difference learning \citep{WeiEtAl2024Lookahead,HuangEtAl2024}. End-to-end mean-field RL learns a policy that maps the operating state and demand features to repositioning shares by origin, trained by backpropagation through the five-minute model \citep{JusupEtAl2026MeanField}. Atomic-PPO decides for small groups of idle vehicles in turn whether to serve, reposition, or stay, using one policy trained by proximal policy optimization \citep{DaiEtAl2025AtomicPPO}. Finally, contextual optimization trains the predictor by decision cost. Per-stage zeroth-order adapts the per-stage training of \citet{GuoEtAl2026}. It trains the predictor on the regret of each stage with a zeroth-order gradient estimate. MCO-serial evaluates each training day serially and uses the exact gradient, and MCO-parallel uses parallel block evaluation (Section~\ref{sec:parallel-trajectory}). Appendix~\ref{app:mco-training} gives their training details. 

\begin{table}[htbp]
  \centering
  \caption{Test Performance of the Compared Methods in the Simulation}
  \label{tab:learning-sim}
  \small
  \setlength{\tabcolsep}{3.5pt}
  \begin{tabular}{lrrrrrr}
    \toprule
    Method & RMSE & \shortstack{Benefit\\(USD)} & \shortstack{Days\\better} & \shortstack{Waiting\\time (min)} & \shortstack{Abandonment\\(\%)} & \shortstack{Completions\\(\%)} \\
    \midrule
    \multicolumn{7}{l}{\emph{Predict-then-optimize}} \\
    MSE & 10.21 & 2.608 & -- & 1.49 & 11.83 & 86.67 \\
    HA & 15.48 & 2.554 $(-2.06\%)$ & 9/20 & 1.54 & 12.09 & 86.41 \\
    STID & 8.88 & 2.624 $(+0.60\%)$ & 14/20 & 1.49 & 11.76 & 86.75 \\
    ST-SSDL & 10.07 & 2.617 $(+0.33\%)$ & 9/20 & 1.49 & 11.80 & 86.71 \\
    MAGE & \textbf{8.64} & 2.611 $(+0.10\%)$ & 11/20 & 1.50 & 11.85 & 86.66 \\
    Oracle & 0.00 & 2.643 $(+1.34\%)$ & 19/20 & 1.49 & 11.62 & 86.89 \\
    \midrule
    \multicolumn{7}{l}{\emph{Optimization under uncertainty}} \\
    SAA & -- & 2.603 $(-0.20\%)$ & 7/20 & 1.51 & 11.89 & 86.62 \\
    CVaR & -- & 2.617 $(+0.33\%)$ & 8/20 & 1.50 & 11.72 & 86.79 \\
    RO & -- & 2.633 $(+0.94\%)$ & 12/20 & 1.50 & 11.52 & 86.99 \\
    \midrule
    \multicolumn{7}{l}{\emph{Heuristic, forecast-free}} \\
    +1 rule & -- & 2.001 $(-23.30\%)$ & 0/20 & 1.70 & 15.98 & 82.52 \\
    Equal distribution & -- & 1.955 $(-25.04\%)$ & 0/20 & 1.71 & 16.22 & 82.30 \\
    Least-squares rebalancing & -- & 1.981 $(-24.03\%)$ & 0/20 & 1.70 & 15.98 & 82.50 \\
    Random walk & -- & 1.697 $(-34.94\%)$ & 0/20 & 1.86 & 18.28 & 80.27 \\
    Lost-request dispatch & -- & 1.753 $(-32.77\%)$ & 0/20 & 1.81 & 17.51 & 80.96 \\
    Real-time rebalancing & -- & 1.407 $(-46.04\%)$ & 0/20 & 1.96 & 19.56 & 79.00 \\
    \midrule
    \multicolumn{7}{l}{\emph{Heuristic, forecast-based}} \\
    Forecast-guided greedy & 10.21 & 2.078 $(-20.32\%)$ & 0/20 & 1.66 & 15.53 & 82.96 \\
    Historical share & 15.48 & 2.086 $(-20.03\%)$ & 0/20 & 1.66 & 15.38 & 83.10 \\
    Target cruise & 10.21 & 1.345 $(-48.42\%)$ & 0/20 & 2.06 & 21.17 & 77.38 \\
    \midrule
    \multicolumn{7}{l}{\emph{Reinforcement learning}} \\
    Learned-value lookahead & 10.21 & 2.684 $(+2.92\%)$ & \textbf{20/20} & 1.47 & 11.47 & 87.04 \\
    End-to-end mean-field RL & -- & 2.011 $(-22.89\%)$ & 0/20 & 1.71 & 16.09 & 82.39 \\
    Atomic-PPO & -- & 1.682 $(-35.49\%)$ & 0/20 & 1.88 & 18.50 & 80.00 \\
    \midrule
    \multicolumn{7}{l}{\emph{Contextual optimization}} \\
    Per-stage zeroth-order & 10.42 & 2.566 $(-1.61\%)$ & 0/20 & 1.52 & 12.21 & 86.29 \\
    MCO-serial & 11.05 & 2.809 $(+7.69\%)$ & \textbf{20/20} & 1.37 & 10.63 & 87.87 \\
    MCO-parallel & 11.17 & \textbf{2.844} $(+9.06\%)$ & \textbf{20/20} & \textbf{1.36} & \textbf{10.38} & \textbf{88.12} \\
    \bottomrule
  \end{tabular}
  \par\smallskip
  \begin{minipage}{\linewidth}\footnotesize\emph{Note.} Parentheses give the relative change in benefit over MSE, computed from unrounded values. Days better counts the test days on which a method has a higher benefit than MSE. Bold marks the best value excluding the oracle.\end{minipage}
\end{table}
MCO performs best in Table~\ref{tab:learning-sim}. It improves the benefit over MSE by 7.69\% to 9.06\% on all 20 test days, with the lowest waiting time and abandonment and the most completions, and MCO-parallel is slightly better than MCO-serial. Its forecasts are less accurate than those of MSE, so training on the decision cost trades accuracy for better decisions. More broadly, the methods that decide with an optimization model stay between $-2.06\%$ and $+2.92\%$ of MSE, whereas the heuristic rules lose 20\% to 48\% and mean-field RL and Atomic-PPO lose 22.89\% and 35.49\%, mainly through more abandonment. Among the optimization-based methods, how the predictor is trained matters more than its accuracy. STID and MAGE have the lowest RMSE but change the benefit by less than 1\%, and per-stage zeroth-order, which trains each stage separately, loses 1.61\%, so the effect of decisions on later states must enter training.

MCO also outperforms the oracle, which gains only 1.34\%, so its gain does not come from accuracy. Exact demand is not the best input because the rolling-horizon policy is itself a heuristic for the multistage problem and the planning LP approximates the simulation with linearized MFD dynamics and aggregate flows. The oracle is therefore no upper bound for this policy. Because MCO trains the predictor on the cost of the executed decisions, it can learn forecasts that partly compensate for both gaps. MCO-parallel forecasts 9.0\% above the demand, most for sparse OD pairs and before the morning peak. The policy then repositions 49\% more vehicles than with MSE, also at midday, and loses 12.3\% fewer requests, whereas the oracle repositions 6\% more and loses 1.8\% fewer.

\subsection{Parallel Block Evaluation}
\label{sec:parallel-results}

This experiment measures how much parallel block evaluation speeds up training and whether its approximate gradient affects the quality of the trained model. We train MCO-serial and MCO-parallel for 30 epochs and evaluate both on the 10 validation days under five-minute execution at the end of each epoch. The validation benefit of both models rises over the epochs, and MCO-parallel is above MCO-serial in 24 of the 29 epochs after the first. The selected checkpoints gain 1.14\% (MCO-parallel, epoch 26) and 1.08\% (MCO-serial, epoch 29) on validation. In the simulation, MCO-parallel gains 9.06\% on the test days and MCO-serial 7.69\% (Table~\ref{tab:learning-sim}). The approximate gradient therefore does not reduce the quality of the trained model. The 30 epochs of training take 19.9 hours for MCO-parallel and 37.1 hours for MCO-serial, and at every epoch end of MCO-serial after the first, the most recent MCO-parallel checkpoint has the higher validation benefit. In the first epoch, MCO-parallel also evaluates every day serially. From the second epoch on, a visit to a training day takes about 75 seconds in MCO-parallel and 144 seconds in MCO-serial, a speedup of 1.91. The difference comes from the forward pass, which takes 38.5 seconds with parallel block evaluation and 104.4 seconds serially. The sequential backward pass takes 36.2 and 39.1 seconds.

\subsection{Operational Insights}
\label{sec:operational-insights}

The results suggest four insights for platform operations. First, repositioning pays off most when the inherited fleet is far from demand. At 06:00, Manhattan South/Core and Brooklyn West/North generate 45.4\% of the requests but hold only 10.3\% of the idle vehicles. MCO-parallel repositions 3,521 vehicles in the first hour against 2,245 with MSE, and 90\% of its reduction in abandoned requests falls before 15:00. Second, en-route pickup makes repositioning affordable. Without pickup, every repositioning trip runs empty, so the planning LP repositions about half as much and loses 6.44\% of its benefit with HA. Third, the value of repositioning lies in its timing and destination rather than its volume. Most heuristic rules drive two to three times the repositioning miles of MSE per completed request, yet they lose 20\% to 48\% of the benefit. MCO-parallel, in contrast, saves about three abandoned requests per additional repositioned vehicle, mostly by moving vehicles from Manhattan South/Core, the two Queens regions, and Staten Island toward the two Brooklyn regions and the Bronx. Fourth, forecasts for repositioning should be judged by the decisions they produce. The most accurate forecasters change the benefit by less than 1\%, and exact demand gains only 1.34\%, whereas MCO-parallel gains 9.06\% with forecasts 9.0\% above demand.

\section{Conclusion}
\label{sec:conclusion}
This paper studied multistage contextual optimization for congestion-aware vehicle repositioning. We formulated a rolling commitment--recourse policy. It fixes repositioning before demand is realized and decides service afterward. A linearized MFD with a penalty on a movement-error bound keeps each stage an LP. We then trained the demand predictor on the decision cost of the whole multistage trajectory. The predictor uses the state left by earlier decisions as context. We derived the exact gradient and an equivalent adjoint that avoids full Jacobians. Finally, we proposed parallel block evaluation to speed up the sequential forward evaluation. It recovers the serial trajectory within a finite number of passes. Its gradient error is bounded under early stopping.

In the New York City case study, evaluated in an agent-based simulation, multistage contextual optimization yields an additional 7.69\% (MCO-serial) and 9.06\% (MCO-parallel) of business benefit over the same predictor trained by mean squared error, on every test day. Both models also outperform the oracle forecast within the same policy. Exact demand therefore does not make this rolling, linearized policy optimal, and MCO learns forecasts that partly compensate for its gaps. For practice, the results suggest concentrating repositioning on the hours when demand moves away from the inherited fleet, letting repositioning vehicles pick up passengers en route, and judging forecasts by the decisions they produce rather than by their accuracy.

Several directions remain open. First, we treat demand as exogenous, but the distribution of vehicles also changes demand: short pickup times attract riders from other modes, so recorded requests depend on the policy that served them. Learning under such decision-dependent demand requires a model of how demand responds to supply. Second, we assume that vehicles follow the repositioning decisions. Human drivers choose where to go, so the platform must design incentives that make its decisions attractive to them. Third, platforms also control prices and matching, which interact with repositioning \citep{WangEtAl2025Dispatch}. Training a predictor on the joint cost of these decisions is a natural extension of multistage contextual optimization.

\FloatBarrier
\bibliographystyle{informs2014trsc}
\bibliography{references}

\clearpage
\appendix
\section{Derivation of the Affine Movement Error Bound}
\label{app:affine-movement-error}

\begin{proof}[Proof of Proposition~\ref{prop:affine-movement-error}]
Write $n$ for class stock and $N$ for regional accumulation, and drop the region index, so that $|\varphi'|\le\bar\varphi'$ and $|\varphi''|\le\bar\varphi''$. Let $f$ denote the affine movement \eqref{eq:affine-period-event}, with deviations $\delta n=n-n^{\mathrm{ref}}$ and $\delta N=N-N^{\mathrm{ref}}$. Expanding the absolute error gives
\begin{equation}
  \begin{aligned}
    |n\varphi(N)-f|
    &=\Bigl|n\varphi(N)-\Bigl[n^{\mathrm{ref}}\varphi(N^{\mathrm{ref}})
      +\varphi(N^{\mathrm{ref}})(n-n^{\mathrm{ref}})
      +n^{\mathrm{ref}}\varphi'(N^{\mathrm{ref}})(N-N^{\mathrm{ref}})\Bigr]\Bigr|\\
    &=\left|n\varphi(N)-n\varphi(N^{\mathrm{ref}})
      -n^{\mathrm{ref}}\varphi'(N^{\mathrm{ref}})\delta N\right|\\
    &=\left|(n^{\mathrm{ref}}+\delta n)
      \bigl[\varphi(N)-\varphi(N^{\mathrm{ref}})\bigr]
      -n^{\mathrm{ref}}\varphi'(N^{\mathrm{ref}})\delta N\right|\\
    &=\left|\delta n\underbrace{\bigl[\varphi(N)-\varphi(N^{\mathrm{ref}})\bigr]}
          _{\text{fraction difference}}
      +n^{\mathrm{ref}}\underbrace{\bigl[\varphi(N)-\varphi(N^{\mathrm{ref}})
          -\varphi'(N^{\mathrm{ref}})\delta N\bigr]}
          _{\text{first-order remainder}}\right|\\
    &\le |\delta n|\underbrace{|\varphi(N)-\varphi(N^{\mathrm{ref}})|}
          _{\le \bar\varphi'|\delta N|}
      +\frac{n^{\mathrm{ref}}}{2}
       \underbrace{|\varphi''(\widetilde N)|}_{\le \bar\varphi''}(\delta N)^2\\
    &\le \bar\varphi'|\delta n|\,|\delta N|
      +\frac{n^{\mathrm{ref}}\bar\varphi''}{2}(\delta N)^2\\
    &\le\frac{\bar\varphi'}{2}\bigl[(\delta n)^2+(\delta N)^2\bigr]
      +\frac{n^{\mathrm{ref}}\bar\varphi''}{2}(\delta N)^2\\
    &=\frac{\bar\varphi'}{2}(\delta n)^2
      +\frac{\bar\varphi'+n^{\mathrm{ref}}\bar\varphi''}{2}(\delta N)^2.
  \end{aligned}
  \label{eq:affine-error-decomposition}
\end{equation}
The equalities substitute the affine movement equation, write $n=n^{\mathrm{ref}}+\delta n$, and collect the fraction difference and first-order remainder. The first inequality uses the triangle inequality and Taylor's theorem, where $\widetilde N$ lies between $N$ and $N^{\mathrm{ref}}$. The mean value theorem bounds the fraction difference by $\bar\varphi'|\delta N|$, and $|\varphi''|\le \bar\varphi''$ bounds the remainder. Finally, Young's inequality $2|\delta n|\,|\delta N|\le(\delta n)^2+(\delta N)^2$ separates the deviations.
\end{proof}

\section{Reference-Trajectory Update Rules}
\label{app:reference-update}

This appendix gives the construction details for \eqref{eq:post-reference-map}--\eqref{eq:initial-reference-map}. All planning periods overlap in the pre-to-post update. In the post-to-pre update, only the new tail period does not overlap. Write $u^\star$ for the source optimal plan and $\mathcal H^\star$ for its set of absolute periods. Shares below are indexed by absolute period.

First, extract allocation shares from this plan as decision preferences. Let $\pi_{ij,\tau}^{m}$ and $\pi_{oj,\tau}^{g}$ be the shares of idle vehicles assigned to ordinary service and repositioning, respectively. Let $\pi_{oij,\tau}^{y}$ be the share of available empty vehicles used for pickup.
A small positive constant $\varsigma>0$ keeps these ratios defined at zero stock:
\begin{equation}
  \begin{aligned}
    \pi_{ij,\tau}^{m}&=\frac{m_{ij,\tau}^{\star}}{e_{i,\tau}^{\star}+\varsigma},
    &&\forall (i,j)\in\Dpax,\ \tau\in\mathcal H^\star,\\
    \pi_{oj,\tau}^{g}&=\frac{g_{oj,\tau}^{\star}}{e_{o,\tau}^{\star}+\varsigma},
    &&\forall (o,j)\in\Doff,\ \tau\in\mathcal H^\star,\\
    \pi_{oij,\tau}^{y}&=\frac{y_{oij,\tau}^{\star}}{\tilde n_{i,\tau}^{a_{oj}^{\mathrm{emp}},\star}+\varsigma},
    &&\forall (o,j)\in\Doff,\ (i,j)\in\Doff,\ i\in\mathcal P_{oj}\setminus\{o,j\},\ \tau\in\mathcal H^\star.
  \end{aligned}
  \label{eq:reference-idle-allocation}
\end{equation}

For a reference constructed at stage $t$, initialize the idle, cohort-backlog, and moving stocks from that stage's initial state $s_t$, and proceed through $\tau=t,\ldots,t+H-1$. At each period, repositioning applies its shares to the available idle stock, \(g_{oj,\tau\mid t}^{\mathrm{ref}}=e_{o,\tau\mid t}^{\mathrm{ref}}\pi_{oj,\tau}^{g}\). In the post reference, first-period repositioning is already committed, so \(g_{oj,t\mid t}^{\mathrm{ref,post}}=g_{oj,t}^{\mathrm{exec}}\).

Ordinary service and pickup must jointly fit the available demand, which is the inherited backlog $q_{ij,\tau\mid t}^{\mathrm{ref}}$ plus period demand $d^{\mathrm{ref}}_{ij,\tau\mid t}$. For the pre reference, $d^{\mathrm{ref}}_{ij,\tau\mid t}=\widehat d_{ij,\tau\mid t}$. For the post reference, it equals realized demand at $\tau=t$ and the forecast at later periods. Let the proposed total service be $\mu_{ij,\tau\mid t}^{\mathrm{prop}}$ and the demand correction fraction be $\upsilon_{ij,\tau\mid t}$:
\begin{equation}
  \begin{gathered}
    \mu_{ij,\tau\mid t}^{\mathrm{prop}}
    =e_{i,\tau\mid t}^{\mathrm{ref}}\pi_{ij,\tau}^{m}
      +\sum_{o:\,(o,j)\in\Doff,\ i\in\mathcal P_{oj}\setminus\{o,j\}}
       \tilde n_{i,\tau\mid t}^{a_{oj}^{\mathrm{emp}},\mathrm{ref}}\pi_{oij,\tau}^{y},
    \quad \forall (i,j)\in\Dpax,\ t\in\mathcal T,\ \tau\in t+\mathcal H,\\
    \upsilon_{ij,\tau\mid t}
    =\begin{cases}
      1,&\mu_{ij,\tau\mid t}^{\mathrm{prop}}=0,\\
      \min\{1,(q_{ij,\tau\mid t}^{\mathrm{ref}}+d^{\mathrm{ref}}_{ij,\tau\mid t})/\mu_{ij,\tau\mid t}^{\mathrm{prop}}\},
         &\mu_{ij,\tau\mid t}^{\mathrm{prop}}>0,
     \end{cases}
    \quad \forall (i,j)\in\Dpax,\ t\in\mathcal T,\ \tau\in t+\mathcal H.
  \end{gathered}
  \label{eq:reference-service-correction}
\end{equation}
The reference ordinary service and pickup are the two proposed terms scaled by this fraction, \(m_{ij,\tau\mid t}^{\mathrm{ref}}=\upsilon_{ij,\tau\mid t}e_{i,\tau\mid t}^{\mathrm{ref}}\pi_{ij,\tau}^{m}\) and \(y_{oij,\tau\mid t}^{\mathrm{ref}}=\upsilon_{ij,\tau\mid t}\tilde n_{i,\tau\mid t}^{a_{oj}^{\mathrm{emp}},\mathrm{ref}}\pi_{oij,\tau}^{y}\). Allocate this service to cohorts oldest first and update backlog and abandonment by \eqref{eq:served-demand}--\eqref{eq:aggregate-backlog}. Vehicles not dispatched remain idle. Compute post-conversion class stocks by \eqref{eq:post-dispatch-stock}--\eqref{eq:post-conversion-loaded} and accumulation by \eqref{eq:euler-accumulation}. Transfers and completions follow \eqref{eq:period-path-flows}, with \(f^a_{k,\tau}\) given by \eqref{eq:nonlinear-period-event}; \eqref{eq:mfd-state-transition} then gives the idle and moving stocks at $\tau+1$, which is the next period's initial state.

The procedure above constructs the reference for periods shared by the source plan and the current problem. The pre problem at stage $t\in\mathcal T\setminus\{0\}$ includes one additional period, $t+H-1$, beyond the preceding post plan. We reuse that plan's allocation shares from its final period $t+H-2$ for the new period. Apply these shares to the new period's propagated stocks and forecast using the rules above. Evaluate movement at that period's own accumulation.

At the first stage, no preceding plan is available, so we obtain one by solving the fixed-fraction LP with $s_0$ and $\widehat d_0$. For each lead $\ell\in\mathcal H$, use this optimum's corresponding quantities in \eqref{eq:reference-idle-allocation}. Starting from $s_0$, apply the repair and nonlinear propagation above under $\widehat d_0$. Thus the fixed-fraction plan supplies the shares, while the endogenous MFD generates the reference stocks.

\section{State-Conditioned Demand Model Details}
\label{app:state-conditioned-predictor}

We specify the factorized demand predictor \(f_\theta\).
For each stage \(t\in\mathcal T\), lead \(\ell\in\mathcal H\), and OD pair \((i,j)\in\Dpax\),
we use the current state \(s_t\) and available exogenous context \(x_t^{\mathrm{exo}}\)
to form four input vectors: global \(x_{t,\ell}^{\mathrm{glob}}\),
origin \(x_{i,t,\ell}^{\mathrm{org}}\), destination \(x_{j,t,\ell}^{\mathrm{dst}}\),
and OD \(x_{ij,t,\ell}^{\mathrm{od}}\).
State features are scaled by the initial fleet size. The global vector contains the total idle and in-motion platform vehicles and the total waiting requests. The origin and destination vectors contain each region's idle vehicles, in-motion platform vehicles, and waiting requests that start or end there. The OD vector contains each pair's waiting requests and in-motion platform vehicles. Each vector also contains exogenous demand features aggregated to its level: demand in recent periods, demand in the target period on the previous day and in the previous week, and the same-weekday historical average for the target period. The global vector also contains the time of day and day of week.
Let \(r_i^{\mathrm{reg}}\) and \(r_\ell^{\mathrm{lead}}\) be learned region and lead embeddings. Encoders \(G_\theta,O_\theta,D_\theta,Q_\theta\), shared across stages, leads, regions, and OD pairs, map the global, origin, destination, and OD input vectors to feature vectors \(\mathbf v_{t,\ell}^{\mathrm{glob}}\), \(\mathbf v_{i,t,\ell}^{\mathrm{org}}\), \(\mathbf v_{j,t,\ell}^{\mathrm{dst}}\), and \(\mathbf v_{ij,t,\ell}^{\mathrm{od}}\). Each input is concatenated with the lead embedding and the embeddings of its regions, and the origin, destination, and OD inputs also with \(\mathbf v_{t,\ell}^{\mathrm{glob}}\).
A scalar score \(\zeta_{ij,t,\ell}\) combines origin, destination, interaction, and pair effects:
\begin{equation}
\begin{aligned}
  \zeta_{ij,t,\ell}
  &={}
  w^0_\ell
  +w_{\mathrm{o}}^\top \mathbf v_{i,t,\ell}^{\mathrm{org}}
  +w_{\mathrm{d}}^\top \mathbf v_{j,t,\ell}^{\mathrm{dst}}
  +w_\ell^{\mathrm{int}}
   \frac{
     (\mathbf v_{i,t,\ell}^{\mathrm{org}})^\top
     \mathbf v_{j,t,\ell}^{\mathrm{dst}}
   }{\sqrt{K^{\mathrm{fac}}}}
  +w_{\mathrm{p}}^\top \mathbf v_{ij,t,\ell}^{\mathrm{od}},\forall (i,j)\in\Dpax,\ t\in\mathcal T,\ \ell\in\mathcal H.
\end{aligned}
  \label{eq:state-conditioned-demand-score}
\end{equation}
Here \(K^{\mathrm{fac}}\) is the dimension of the origin and destination feature vectors. The parameter vector \(\theta\) collects the encoder weights, the embeddings \(r^{\mathrm{reg}}\) and \(r^{\mathrm{lead}}\), and the scoring weights \(w^0_\ell\), \(w_{\mathrm{o}}\), \(w_{\mathrm{d}}\), \(w_\ell^{\mathrm{int}}\), and \(w_{\mathrm{p}}\).
The forecast \(\widehat d_{ij,t+\ell\mid t}=\log(1+\exp(\zeta_{ij,t,\ell}))\) is therefore nonnegative.

\section{Stage Sensitivities}
\label{app:stage-sensitivity}

Differentiate the post plan \eqref{eq:commitment-recourse-action} with the Jacobian \eqref{eq:post-reallocation-jacobian}. Let \(\mathsf J_{u,x,t}^{\mathrm{post}}\) denote its partial Jacobian with respect to the input \(x\in\{s,\widehat d^{\mathrm{fut}},g,\boldsymbol n\}\). Then
substitute the pre-solution and commitment gradients from
\eqref{eq:pre-commitment-sensitivity}, together with the reference gradients
\eqref{eq:pre-reference-sensitivity} and \eqref{eq:post-reference-sensitivity}.
Finally, substitute the predictor gradient
\eqref{eq:state-conditioned-demand-sensitivity} and use
\(s_t=\Pi_su_{t-1}^{\mathrm{post},\star}\) for later stages. This gives the
three terms in \eqref{eq:state-conditioned-post-solution-sensitivity}:
the direct prediction effect, the effect through the preceding plan, and the effect through state-conditioned
prediction. We expand their coefficients below.

Changes in the pre optimum reach the post plan through the commitment and the post reference:
\begin{equation}
  \mathsf J_{u,u,t}^{\mathrm{post,pre}}
  :=\mathsf J_{u,g,t}^{\mathrm{post}}\Pi_g
   +\mathsf J_{u,\boldsymbol n,t}^{\mathrm{post}}\mathsf J_{\boldsymbol n,u,t}^{\mathrm{post}},
  \qquad\forall t\in\mathcal T.
  \label{eq:appendix-post-chain}
\end{equation}
The first term in \eqref{eq:state-conditioned-post-solution-sensitivity} uses the forecast sensitivity. A forecast changes the post plan directly and through the post reference, and it changes the pre optimum directly and through the pre reference:
\begin{equation}
  \mathsf J_{u,\widehat d,t}^{\mathrm{stage}}
  =\bigl(\mathsf J_{u,\widehat d^{\mathrm{fut}},t}^{\mathrm{post}}
   +\mathsf J_{u,\boldsymbol n,t}^{\mathrm{post}}\mathsf J_{\boldsymbol n,\widehat d^{\mathrm{fut}},t}^{\mathrm{post}}\bigr)\Pi_{\mathrm{fut}}
   +\mathsf J_{u,u,t}^{\mathrm{post,pre}}
   \bigl(\mathsf J_{u,\widehat d,t}^{\mathrm{pre}}
   +\mathsf J_{u,\boldsymbol n,t}^{\mathrm{pre}}\mathsf J_{\boldsymbol n,\widehat d,t}^{\mathrm{pre}}\bigr),
  \qquad\forall t\in\mathcal T.
  \label{eq:appendix-stage-coefficients}
\end{equation}
The second term holds the forecast fixed and follows the preceding plan. The state enters both LPs directly and through both references:
\begin{equation}
  \mathsf J_{u,s,t}^{\mathrm{stage}}
  =\mathsf J_{u,s,t}^{\mathrm{post}}
   +\mathsf J_{u,\boldsymbol n,t}^{\mathrm{post}}\mathsf J_{\boldsymbol n,s,t}^{\mathrm{post}}
   +\mathsf J_{u,u,t}^{\mathrm{post,pre}}
   \bigl(\mathsf J_{u,s,t}^{\mathrm{pre}}
   +\mathsf J_{u,\boldsymbol n,t}^{\mathrm{pre}}\mathsf J_{\boldsymbol n,s,t}^{\mathrm{pre}}\bigr),
  \qquad\forall t\in\mathcal T.
  \label{eq:appendix-stage-state-coefficient}
\end{equation}
The preceding plan acts through the inherited state and through the pre reference, which gives the coefficient of the second term:
\begin{equation}
  \mathsf J_{u,u,t}^{\mathrm{stage}}
  =\mathsf J_{u,s,t}^{\mathrm{stage}}\Pi_s
   +\mathsf J_{u,u,t}^{\mathrm{post,pre}}\mathsf J_{u,\boldsymbol n,t}^{\mathrm{pre}}\mathsf J_{\boldsymbol n,u,t}^{\mathrm{pre}}
  =\mathsf J_{u,s,t}^{\mathrm{stage}}\Pi_s+\mathsf J_{u,u,t}^{\mathrm{ref}},
  \qquad\forall t\in\mathcal T\setminus\{0\}.
  \label{eq:appendix-inherited-stage-coefficient}
\end{equation}
The third term follows parameter changes through earlier decisions into the inherited state and then through prediction into the current post optimum, with coefficient \(\mathsf J_{u,\widehat d,t}^{\mathrm{stage}}\mathsf J_{\widehat d,s,t}\mathsf J_{s,\theta,t}\) for \(t\in\mathcal T\setminus\{0\}\).

\section{Proofs for Parallel Block Evaluation}
\label{app:parallel-trajectory-proof}

\begin{proof}[Proof of Proposition~\ref{prop:trajectory-recovery}(i)]
A block that starts from the serial plan solves the same LPs as serial evaluation. By the condition in part~(i), it then selects the same solutions, so it reproduces serial evaluation and outputs the serial plan at its end. Block \(0\) starts from the given state \(s_0\). Pass \(1\) therefore reproduces serial evaluation in block \(0\). By the update \eqref{eq:trajectory-round-transition}, the output of block \(0\) becomes the entry of block \(1\) in pass \(2\). This entry is the serial plan, so pass \(2\) reproduces serial evaluation in block \(1\). Its output becomes the entry of block \(2\) in pass \(3\), and so on. By induction, pass \(\kappa\) reproduces serial evaluation in block \(\kappa-1\). Later passes do not solve this block again, so it keeps these serial records. After pass \(M\), all \(M\) blocks hold serial records, and the final records reproduce the serial trajectory.
\end{proof}

\begin{proof}[Proof of Proposition~\ref{prop:trajectory-recovery}(ii)]
Let \(e_j^{(\kappa)}=\|\bar u_j^{(\kappa)}-u_{\bar t_j-1}^{\mathrm{ser}}\|\) be the entry error of block \(j\) in pass \(\kappa\). For \(\kappa=1\), the bound holds trivially because the product is empty and equals one. Fix \(\kappa\in\{2,\ldots,K\}\) and \(j\in\{\kappa,\ldots,M-1\}\). By the update \eqref{eq:trajectory-round-transition}, the entry of block \(j\) in pass \(\kappa\) is the output of block \(j-1\) in pass \(\kappa-1\). The serial plan at the same stage is the output of the same block map at the serial entry. Hence
\begin{equation}
  \begin{aligned}
    e_j^{(\kappa)}
    &=\bigl\|\mathcal F_{j-1}\bigl(\bar u_{j-1}^{(\kappa-1)};\theta\bigr)
       -\mathcal F_{j-1}\bigl(u_{\bar t_{j-1}-1}^{\mathrm{ser}};\theta\bigr)\bigr\|
    \le\Bigl(\prod_{t\in\mathcal S_{j-1}}L_t\Bigr)e_{j-1}^{(\kappa-1)}\\
    &\le\Bigl(\prod_{t\in\mathcal S_{j-2}\cup\mathcal S_{j-1}}L_t\Bigr)e_{j-2}^{(\kappa-2)}
    \le\cdots
    \le\Bigl(\prod_{t=\bar t_{j-\kappa+1}}^{\bar t_j-1}L_t\Bigr)e_{j-\kappa+1}^{(1)}.
  \end{aligned}
  \label{eq:trajectory-pass-recursion}
\end{equation}
Each inequality holds because a block map composes Lipschitz stage maps, so it is Lipschitz with the product of their constants. Each step moves one block upstream and one pass back, until it reaches the candidate entry of pass \(1\). This is \eqref{eq:trajectory-pass-error}.
\end{proof}

\begin{proof}[Proof of Proposition~\ref{prop:trajectory-recovery}(iii)]
We write the plans of a day as \(X=(u_0,\ldots,u_{T-1})\). The serial plans give \(X^{\mathrm{ser}}\). The final records of pass \(K\) give \(\check X\), whose component \(\check u_t\) is the plan that stage \(t+1\) reads. At the end of block \(j-1\), the entry \(\bar u_j^{(K)}\) differs from the block output by the residual \(r_{j-1}^{(K)}\). Moving all terms of the stage equations to one side, the final records therefore satisfy the residual system
\begin{equation*}
  R(\check X;\theta)=r^{(K)},\qquad
  R(X;\theta)
  =\bigl(u_0-F_0(\theta),\ u_1-F_1(u_0;\theta),\ \ldots,\ u_{T-1}-F_{T-1}(u_{T-2};\theta)\bigr),
\end{equation*}
The backward pass of Section~\ref{sec:inexact-trajectory-training} holds the residuals fixed. Differentiating this system with respect to \(\theta\) at \(\check X\) gives
\begin{equation*}
  \frac{\mathrm d\check X}{\mathrm d\theta}
  =-\Bigl(\frac{\partial R}{\partial X}\Bigr)^{-1}\frac{\partial R}{\partial\theta}.
\end{equation*}
Write the loss of sample \(\omega\) as a function \(\mathcal L^\omega(X;\theta)\) of the stacked plans, so that \(\mathcal L^\omega(\theta)=\mathcal L^\omega(X^{\mathrm{ser}};\theta)\). The chain rule then gives the training direction as \(\widetilde\nabla=g(\check X)\), where for any stacked plans \(X\)
\begin{equation}
  g(X)=\nabla_\theta\mathcal L^\omega(X;\theta)
   -\Bigl(\frac{\partial R}{\partial\theta}\Bigr)^{\!\top}
    \Bigl(\frac{\partial R}{\partial X}\Bigr)^{\!-\top}\nabla_X\mathcal L^\omega(X;\theta),
  \label{eq:trajectory-stacked-adjoint}
\end{equation}
The error of the training direction is therefore the change of \(g\) between \(\check X\) and \(X^{\mathrm{ser}}\). We use the infinity norm \(\|X\|=\max_{t\in\mathcal T}\|u_t\|\). By the conditions of part~(iii), every term of \(g\) is bounded and Lipschitz near \(X^{\mathrm{ser}}\). Sums and products of such terms remain Lipschitz, so \(g\) is Lipschitz with some constant \(\Lambda_\nabla\):
\begin{equation}
  \bigl\|\widetilde\nabla-\nabla_\theta\mathcal L^\omega(\theta)\bigr\|
  =\|g(\check X)-g(X^{\mathrm{ser}})\|
  \le\Lambda_\nabla\|\check X-X^{\mathrm{ser}}\|.
  \label{eq:trajectory-gradient-error-bound}
\end{equation}

The distance between the trajectories comes only from the block entries. Block \(0\) has serial records. Block \(j\in\mathcal M\setminus\{0\}\) starts from \(\bar u_j^{(K)}\), and, as in part~(ii), the error of its stage \(t\) is at most \(\prod_{t'=\bar t_j}^{t}L_{t'}\) times this entry error. With \(\Lambda_L\) the largest of one and these products,
\begin{equation*}
  \|\check X-X^{\mathrm{ser}}\|
  \le\Lambda_L\max_{j\in\mathcal M\setminus\{0\}}\bigl\|\bar u_j^{(K)}-u_{\bar t_j-1}^{\mathrm{ser}}\bigr\|.
\end{equation*}
Inserting this into \eqref{eq:trajectory-gradient-error-bound} gives \eqref{eq:trajectory-gradient-error} with \(\Lambda=\Lambda_\nabla\Lambda_L\).
\end{proof}

\section{Experimental Details}
\label{app:additional-results}

\subsection{Data}
\label{app:data}

The statistics below cover the 20 test days, 14 weekdays and 6 weekend days, between 06:00 and 21:00.

\emph{Demand.} A test day has 493,625 requests on average, and the daily totals range from about 435,000 to 618,000. On weekdays, requests per five minutes reach a morning peak of 3,639 at 08:40 and an evening peak of 3,383 at 18:00. Weekends have no sharp morning peak and reach 3,607 at 20:00. Manhattan South/Core and Brooklyn West/North generate 45.4\% of the requests (Figure~\ref{fig:study-area}a).

\emph{Fleet and initial states.} The fleet of a day is set from the peak number of vehicles on trips on the same weekday in the previous eight weeks, and it is at least the number of vehicles on trips at 04:00. It averages 13,290 vehicles on the test days. At 04:00, the vehicles on trips are taken from the trip records and placed along their routes, no request is waiting, and the remaining vehicles are spread evenly over the regions. The five-minute model then runs without repositioning until 06:00 and gives the initial state of the day. At 06:00, the fleet is far from the demand: Manhattan South/Core and Brooklyn West/North hold only 10.3\% of the idle vehicles, whereas the two Queens regions hold 50.2\% (Figure~\ref{fig:study-area}b). This mismatch makes early repositioning valuable.

\emph{Traffic.} Each region follows the speed MFD \(v_k(N)=v^{\mathrm f}_k(1-N/N_k^{\mathrm{jam}})^2\), so its production is \(P_k(N)=Nv_k(N)\) and its critical accumulation is \(N_k^{\mathrm{jam}}/3\). The free-flow speed \(v^{\mathrm f}_k\) and \(N_k^{\mathrm{jam}}\) are fitted to the regional speeds and accumulations of March 2026 in 15-minute intervals (Table~\ref{tab:mfd-parameters}). The congested branch is rarely observed, so \(N_k^{\mathrm{jam}}\) is a fitted parameter rather than an observed jam accumulation. The background traffic \(N^{\mathrm{bg}}_{k,\tau}\), all vehicles other than ride-hailing, is reconstructed per region and 15-minute interval from county vehicle miles traveled, road traffic counts with their time profiles, and taxi and transit activity, and it is held constant within each 15 minutes. The platform accounts for 7.1\% of the total accumulation on average.

\emph{Planning model parameters.} The mean trip length \(\bar\ell_k\) is the mean distance that a trip travels in region \(k\) per regional movement, estimated from the March trip records along their region paths. With \(\Delta=5\) minutes, the movement fraction \(\varphi_k\) is largest at zero accumulation, where it equals \(\Delta v^{\mathrm f}_k/\bar\ell_k\). This value lies between 0.51 and 0.79, so \(\varphi_k<1\) holds in every region. The envelope \(\Psi\) uses the points \(\xi_h\) at 0, \(\pm1\), \(\pm5\), \(\pm20\), \(\pm100\), \(\pm500\), \(\pm2{,}000\), and \(\pm10{,}000\) vehicles and at the endpoints \(\pm\max_kN_k^{\mathrm{jam}}\), which give \(S=16\) segments.

\begin{table}[htbp]
  \centering
  \caption{Regional MFD Parameters}
  \label{tab:mfd-parameters}
  \small
  \begin{tabular}{lrrrr}
    \toprule
    Region & \(\bar\ell_k\) (mi) & \(v^{\mathrm f}_k\) (mph) & \(N_k^{\mathrm{jam}}\) (veh) & Largest \(\varphi_k\) \\
    \midrule
    Manhattan South/Core & 2.805 & 20.46 & 83{,}026 & 0.608 \\
    Manhattan North & 2.546 & 24.27 & 49{,}963 & 0.794 \\
    Bronx & 2.927 & 20.56 & 204{,}463 & 0.585 \\
    Brooklyn West/North & 2.796 & 18.52 & 170{,}211 & 0.552 \\
    Brooklyn East/South & 2.279 & 17.41 & 126{,}993 & 0.636 \\
    Queens West & 3.521 & 23.35 & 90{,}368 & 0.553 \\
    Queens East & 3.975 & 24.43 & 290{,}622 & 0.512 \\
    Staten Island & 3.735 & 22.99 & 143{,}238 & 0.513 \\
    \bottomrule
  \end{tabular}
  \par\smallskip
\end{table}

\subsection{Agent-Based Simulation}
\label{app:simulation}

The simulation runs from 06:00 to 21:00 and is not reset between stages. Each request appears at its recorded time at a point in its TLC taxi zone and abandons after a patience drawn uniformly between 5 and 10 minutes. The 262 zones are nested in the eight regions. A trip follows the region path of its OD pair, the same as in the planning model, along the shortest polyline through the transitions between consecutive regions, which are shared land borders, bridges, and tunnels. Passenger trips cover their recorded miles. Pickup legs and repositioning trips cover the straight-line and polyline distances, multiplied by detour factors fitted on the training days. Vehicles move in 15-second steps at the MFD speed of their current region, which depends on the accumulation of background traffic and fleet vehicles in that region. The zones are small, 1.2 square miles on average, so we do not add a within-zone control such as the coverage-based rebalancing of \citet{ZhuEtAl2024Coverage}.

Every five minutes, the simulation receives the service, repositioning, and en-route pickup numbers of a policy and carries them out as hard limits in the order of the planning model. Fractional numbers are rounded by error diffusion, so the executed totals stay within half a unit of the planned totals over the day. Repositioned vehicles leave the zones of the origin region with the largest surplus of idle vehicles for the zones of the destination region with the largest shortfall. Every 60 seconds, en-route pickups come first: repositioning vehicles in a region are paired, closest pairs first, with waiting requests there whose destination is their target region. Service follows: waiting requests within the service number of their OD pair take, oldest first, the nearest idle vehicle of their region.

\subsection{Implementation of the Compared Methods}
\label{app:experiment-details}

All methods are trained on the 31 March days and observe the same information. In every period, the service is set at the start of the period from the waiting requests and the requests arriving in the period, and the simulation executes it as a hard limit. The LP-based methods commit their repositioning before the demand of the period is realized, whereas the rules reposition after serving when this demand is known. For the forecasters trained for accuracy, the learning rate is selected by the validation error of the forecast. All other hyperparameters and checkpoints are selected by the business benefit on the 10 validation days under five-minute execution. STID, ST-SSDL, and MAGE use the settings of their papers on the same data.

\emph{Optimization under uncertainty.} At each stage, the five days of January to March whose OD demand in the last hour is closest to the current one provide five scenarios. The pre LP shares the current repositioning across scenarios, and the post LP shares the current service and pickup and plans later periods per scenario. CVaR averages the two worst of the five scenario costs.

\emph{Heuristic rules.} Service favors the newest waiting requests, which raised the validation benefit of every rule. Each rule then sets a target number of idle vehicles for every region and moves a share of the gap to this target, with the share selected on the validation days. The +1 rule targets where recent requests started, equal distribution spreads the vehicles evenly, least-squares rebalancing matches them to the waiting requests, random walk spreads part of them over the adjacent regions, lost-request dispatch targets where requests were abandoned in the previous period, and real-time rebalancing evens them out together with the vehicles about to arrive. Forecast-guided greedy dispatch and historical share distribute them in proportion to the MSE and HA forecasts, and target cruise sends them to the nearby region with the highest forecast. Finally, repositioning vehicles pick up the remaining requests on their path whose destination is their target region.

\emph{Reinforcement learning.} In learned-value lookahead, the value of a vehicle in a region at a given time is the completion reward of its next trip plus the discounted value at the destination and arrival time of that trip. It is learned by temporal-difference learning on rollouts of the training days and improved by one policy-iteration step. End-to-end mean-field RL serves by the rule of the heuristic methods and distributes the remaining idle vehicles with a multilayer perceptron, trained by backpropagating the business cost of whole days through the five-minute model. Atomic-PPO first imitates the decisions of MSE and is then trained by proximal policy optimization. \citet{DaiEtAl2025AtomicPPO} designed it for 10 regions and about 300 vehicles. Our fleet has about 13,000 vehicles, so each decision unit is a group of vehicles.

\emph{Contextual optimization.} Per-stage zeroth-order is not a reproduction of \citet{GuoEtAl2026}, which trains STID forecasts against a rebalancing MILP, but keeps its two elements: a decision loss for each stage and a perturbation-based gradient estimate. With the state of a stage fixed, the forecast determines the repositioning, and the loss is the cost of the post LP under the realized demand with this repositioning. The gradient is estimated from one Gaussian perturbation of the forecast per stage. The method starts from MSE and uses the optimizer, prediction weight \(\alpha\), and number of epochs of MCO, so that the two differ only in the training signal.

\subsection{Training of MCO}
\label{app:mco-training}

The MSE model has the architecture of Appendix~\ref{app:state-conditioned-predictor} with the weights of the state inputs fixed at zero. MCO-serial and MCO-parallel start from it and learn the state weights during training. In the objective \eqref{eq:joint-contextual-training-objective}, \(\ell_{\mathrm{pred},t}\) is the mean squared forecast error over the \(6\times64\) horizon entries, normalized by the squared mean March OD demand per five-minute period. We set \(\alpha=1{,}389.54\), so that the weighted prediction loss of the MSE model equals 10\% of the waiting, abandonment, and repositioning costs on two reference training days. A visit to a training day evaluates the day forward, computes the training direction with the adjoint, and makes one Adam update with learning rate \(3\times10^{-4}\). Both models are trained for 30 epochs, each visiting the 31 training days in the same random order. MCO-parallel uses \(M=16\) blocks, \(K=2\) passes, and eight workers. Its first visit to each day is evaluated serially, and later visits take their block entries from the previous visit to the same day.

The gradient is exact only in a neighborhood in which every LP keeps its optimal basis and every rule stays on its branch. Near a basis change, this neighborhood is very small, and the backward recursion over up to 180 stages can return a gradient orders of magnitude larger than usual. An update along such a gradient can raise the cost, and it inflates the step-size estimates of Adam for later updates. We therefore skip an update when its gradient is much larger than the recent ones. MCO-serial skips 35 of its 930 visits, and MCO-parallel skips 152.

\end{document}